\documentclass[preprint,12pt]{elsarticle}

\usepackage{amsmath,amsfonts}
\usepackage{algorithmic}
\usepackage{array}
\usepackage[tight,footnotesize]{subfigure}
\usepackage{textcomp}
\usepackage{stfloats}
\usepackage{url}
\usepackage{verbatim}
\usepackage{graphicx}
\def\BibTeX{{\rm B\kern-.05em{\sc i\kern-.025em b}\kern-.08em
    T\kern-.1667em\lower.7ex\hbox{E}\kern-.125emX}}
\usepackage{balance}
\usepackage{booktabs}
\usepackage{caption}
\usepackage{makecell}
\usepackage{adjustbox}
\usepackage{wrapfig}
\usepackage[linesnumbered,ruled]{algorithm2e}
\usepackage{amssymb}
\usepackage{lineno}
\usepackage{float}
\usepackage{algorithmic}
\usepackage[tight,footnotesize]{subfigure}
\usepackage{booktabs}
\usepackage{caption}
\usepackage{makecell}
\usepackage{adjustbox}
\usepackage{wrapfig}
\usepackage{xcolor}
\usepackage[linesnumbered,ruled]{algorithm2e}

\newcommand{\cmark}{\ding{51}}%
\newcommand{\xmark}{\ding{55}}%
\newcommand{\pmark}{\ding{108}}       

\journal{Journal Name}

\begin{document}
\sloppy
\setlength{\parskip}{0pt}

\begin{frontmatter}


\title{Towards Responsible and Explainable AI Agents with Consensus-Driven Reasoning}



\author[label1]{Eranga Bandara}
\ead{cmedawer@odu.edu}

\author[label2]{Tharaka Hewa}
\ead{tharaka.hewa@oulu.fi}

\author[label1]{Ross Gore}
\ead{rgore@odu.edu}

\author[label1]{Sachin Shetty}
\ead{sshetty@odu.edu}

\author[label1]{Ravi Mukkamala}
\ead{mukka@odu.edu}

\author[label1]{Peter Foytik}
\ead{pfoytik@odu.edu}

\author[label1]{Safdar H. Bouk}
\ead{sbouk@odu.edu}

\author[label3]{Abdul Rahman}
\ead{abdulrahman@deloitte.com}

\author[label4]{Xueping Liang}
\ead{xuliang@fiu.edu}

\author[label8]{Amin Hass}
\ead{amin.hassanzadeh@accenture.com}

\author[label7]{Sachini Rajapakse}
\ead{sachini.rajapakse@iciclelabs.ai}

\author[label5]{Ng Wee Keong}
\ead{awkng@ntu.edu.sg}

\author[label6]{Kasun De Zoysa}
\ead{kasun@ucsc.cmb.ac.lk}

\author[label9]{Aruna Withanage}
\ead{aruna@effectz.ai}
\author[label9]{Nilaan Loganathan}
\ead{nilaan@effectz.ai}

\address[label1]{Old Dominion University, Norfolk, VA, USA}
\address[label2]{Center for Wireless Communications, University of Oulu, Finland}
\address[label3]{Deloitte \& Touche LLP, USA}
\address[label4]{Florida International University, USA}
\address[label5]{Nanyang Technological University, Singapore}
\address[label6]{University of Colombo, Sri Lanka}
\address[label7]{IcicleLabs.AI}
\address[label8]{Accenture Technology Labs, Arlington, VA, USA}
\address[label9]{Effectz.AI}

\begin{abstract}

Agentic AI represents a major shift in how autonomous systems reason, plan, and execute multi-step tasks through the coordination of Large Language Models (LLMs), Vision–Language Models (VLMs), tools, and external services. While these systems enable powerful new capabilities, increasing autonomy introduces critical challenges related to explainability, accountability, robustness, and governance, especially when agent outputs influence downstream actions or decisions. Existing agentic AI implementations often emphasize functionality and scalability, yet provide limited mechanisms for understanding decision rationale or enforcing responsibility across agent interactions. This paper presents a Responsible(RAI) and Explainable(XAI) AI Agent Architecture for production-grade agentic workflows based on multi-model consensus and reasoning-layer governance. In the proposed design, a consortium of heterogeneous LLM and VLM agents independently generates candidate outputs from a shared input context, explicitly exposing uncertainty, disagreement, and alternative interpretations. A dedicated reasoning agent then performs structured consolidation across these outputs, enforcing safety and policy constraints, mitigating hallucinations and bias, and producing auditable, evidence-backed decisions. Explainability is achieved through explicit cross-model comparison and preserved intermediate outputs, while responsibility is enforced through centralized reasoning-layer control and agent-level constraints. We evaluate the architecture across multiple real-world agentic AI workflows, demonstrating that consensus-driven reasoning improves robustness, transparency, and operational trust across diverse application domains. This work provides practical guidance for designing agentic AI systems that are autonomous and scalable, yet responsible and explainable by construction.

\end{abstract}

\begin{keyword}
Agentic AI \sep Agentic AI Workflow \sep Responsible AI \sep Explainable AI \sep LLM \sep Model Context Protocol
\end{keyword}

\end{frontmatter}

\section{Introduction}

The rapid advancement of Large Language Models (LLMs)~\cite{llm, gpt-llm}, Vision–Language Models (VLMs)~\cite{vision-language-model, pixtral, qwen2}, and tool-augmented reasoning has accelerated the adoption of agentic AI systems composed of autonomous agents capable of reasoning, planning, invoking tools, and executing multi-step workflows without continuous human supervision~\cite{agentic-ai, agentsway}. Unlike traditional prompt–response interactions, agentic AI systems decompose complex tasks across multiple specialized agents, each operating with distinct models, tools, and contextual memory, and coordinate them through orchestration logic to form dynamic, goal-driven pipelines. These agentic workflows have enabled powerful real-world applications ranging from content generation and analytics to regulatory compliance, cybersecurity automation, and multimodal media synthesis~\cite{agentic-workflow-practicle-guide}. However, as agentic AI systems transition from experimental prototypes to production deployments, their increasing autonomy introduces fundamental challenges related to responsibility and explainability~\cite{llm-explainability, responsible-ai}. Agent decisions are often derived from complex interactions among multiple models, tools, and intermediate states, making it difficult to understand why a particular action was taken, which model influenced the outcome, or how uncertainty and disagreement were resolved. Single-model reasoning pipelines, still common in many agent implementations, are especially vulnerable to hallucinations, reasoning drift, prompt injection, and silent failure modes, producing outputs that are difficult to audit or justify in high-stakes environments~\cite{agent-survey, llm-attack}. In such systems, incorrect or unsafe decisions can propagate across agent boundaries and downstream tools, amplifying risk rather than containing it. Explainable AI (XAI) and Responsible AI (RAI) have therefore emerged as critical requirements for agentic systems, yet they are often conflated or addressed only superficially~\cite{xai-llm, responsible-ai, agentic-workflow-practicle-guide}. Explainability refers to the ability to understand, inspect, and reason about why an agent produced a given output, exposing intermediate reasoning steps, alternative interpretations, uncertainty, and model disagreement. In contrast, responsibility concerns how agent behavior is governed, ensuring safety, robustness, policy compliance, bias mitigation, accountability, and auditability across autonomous decision-making processes. Although explainability supports human understanding and trust, responsibility ensures that agentic systems behave within acceptable operational and ethical boundaries. In practice, many agentic workflows deployed offer neither: decisions are opaque, governance is implicit, and accountability is difficult to establish once agents act autonomously~\cite{agent-survey}.


Existing approaches to Responsible AI in agentic systems often focus on static safeguards such as prompt constraints, rule-based filters, or post-hoc logging, while explainability is frequently reduced to natural-language justifications generated by the same model that produced the decision~\cite{llm-explainability, xai-llm}. These approaches do not address the deeper architectural issue: responsibility and explainability cannot reliably emerge from a single-model pipeline that lacks independent perspectives, explicit governance, and structured decision synthesis. As agentic systems scale in complexity, a more principled design approach is required, one that treats responsibility and explainability as architectural properties, not afterthoughts~\cite{reconcile}.

In this paper, we propose an architectural framework for building Responsible and Explainable AI agents grounded in two core principles: \textbf{multi-model consensus} and \textbf{reasoning-based governance}~\cite{agentic-workflow-practicle-guide}. Our approach integrates a consortium of heterogeneous LLMs and VLMs, each independently generating candidate outputs for a given task from a shared input context, with a dedicated reasoning agent that evaluates, reconciles, and governs these outputs. The LLM/VLM consortium could show disagreement, uncertainty, and alternative interpretations among models, forming the basis for explainability through explicit cross-model comparison. The reasoning agent, implemented using a specialized reasoning-focused LLM, serves as a governance layer that enforces safety and policy constraints, filters unsafe or speculative content, resolves conflicts, and synthesizes a final evidence-backed decision traceable to its contributing sources~\cite{reasoning-llms, llm-reasoning}. Together, these components enable agentic workflows that are explainable by design and responsible by construction. We evaluate the framework in multiple agentic AI workflows, including the generation of news podcasts, neuromuscular reflex analysis, detection of dental conditions and gingivitis, psychiatric diagnosis, and classification of RF signals, demonstrating how consensus-driven reasoning-based agent orchestration improves robustness, transparency, and operational trust in diverse domains of high-impact applications~\cite{agentic-workflow-practicle-guide, nurolense, deep-psychiatric}. The contributions of this paper are as follows:

\begin{enumerate}
    \item \textbf{A clear architectural separation between explainability and responsibility in agentic AI systems.}  
    We formalize the distinct roles of multi-model consensus for explainability and reasoning-layer governance for responsibility, and show how they jointly address key failure modes in autonomous agents.

    \item \textbf{A practical framework for responsible and explainable agentic workflows.}  
    We present an implementable architecture combining LLM/VLM consortia with a reasoning agent, suitable for real-world production-grade agentic workflows.


    \item \textbf{An evaluation across multiple responsible and explainable AI agent use cases.}  
    We analyze how the proposed approach improves transparency, robustness, and accountability across diverse agentic decision-making scenarios.
\end{enumerate}

The remainder of this paper is organized as follows. Section 2 introduces the Responsible and Explainable Agent Architecture, describing the core design principles, architectural components, and governance mechanisms that underpin the proposed approach. Section 3 presents the implementation and evaluation of this architecture through five representative use cases of the agentic AI workflow: creation of news podcasts, Neuromuscular Reflex Analysis, detection of Tooth-Level Condition and gingivitis, Psychiatric diagnosis, and RF Signal Classification. These use cases demonstrate how the architecture supports multi-agent orchestration, multimodal data processing, reasoning-based consolidation, and Responsible-AI–aligned decision-making across diverse real-world domains. Section 4 reviews related work in agentic AI, multi-model systems, and responsible and explainable AI frameworks. Finally, Section 5 concludes the paper by summarizing key insights and findings and outlining directions for future research, architectural refinement, and the broader adoption of robust, explainable, and trustworthy agentic AI systems.

\section{Consensus Driven Reasoning Architecture}

This section describes the proposed system architecture that enables responsible and explainable AI agents to be implemented within agentic workflows using a combination of multi-model consensus and reasoning-layer governance. Rather than treating explainability and responsibility as post-hoc properties, our approach embeds them directly into the agent architecture, interaction patterns, and decision lifecycle. The architecture explicitly separates decision generation from decision governance, enabling autonomous behavior while maintaining transparency, accountability, and operational control.

\subsection{Design Requirements for Responsibility and Explainability}

In autonomous agentic systems, responsibility and explainability impose distinct but complementary requirements. Explainable AI (XAI) requires that agent decisions be interpretable, inspectable, and attributable, exposing uncertainty, alternative interpretations, and the rationale behind outputs~\cite{xai, llm-explainability}. Responsible AI (RAI), on the contrary, requires that agent behavior be governed, constrained, and auditable, ensuring safety, robustness, policy compliance, and accountability across autonomous decision-making processes~\cite{responsible-ai, agentsway}.

\begin{figure}[H]
\centering{}
\includegraphics[width=5.4in]{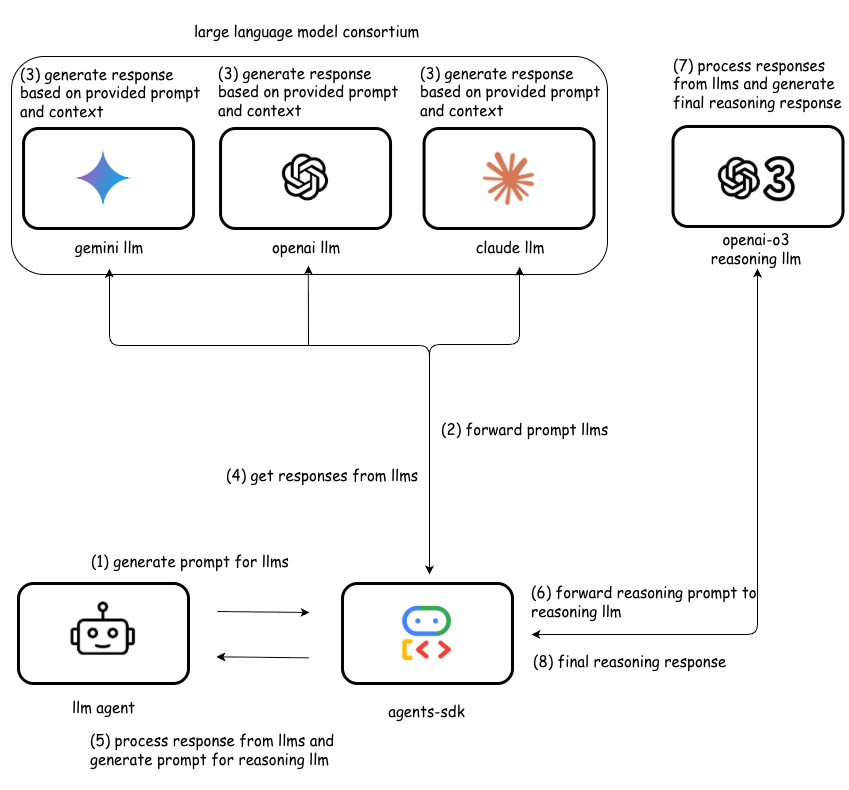}
\vspace{-0.1in}
\DeclareGraphicsExtensions.
\caption{Integration flow of the LLM/VLM consortium with the reasoning-layer governance agent.}
\label{llm-consortium1}
\end{figure}

Single-model agent pipelines do not reliably satisfy these requirements. They provide limited visibility into decision alternatives, collapse uncertainty into a single output, and lack explicit governance mechanisms for filtering unsafe or speculative behavior. As agent autonomy increases, these limitations can lead to hallucinated content, biased decisions, silent failures, and untraceable actions, particularly when agents invoke external tools or trigger downstream effects~\cite{llm-attack, agentic-ai-challenges}. Addressing these challenges requires architectural mechanisms that (i) expose multiple independent perspectives for explainability and (ii) enforce centralized reasoning and control for responsibility. Figure~\ref{llm-consortium1} illustrates the high-level integration of the LLM/VLM consortium with the governance agent of the reasoning-layer.

\subsection{LLM/VLM Consortium for Explainable Agent Decisions}

To support explainability, the proposed architecture employs a consortium of heterogeneous LLMs and VLMs (e.g., GPT, Gemini, Claude, Llama, Pixtral, Qwen)~\cite{llama-3, pixtral, qwen2}. Each model operates as an independent agent that generates a candidate output for a given task. For each execution step, the orchestration layer constructs a single canonical prompt and a shared input context, which are sent unchanged to all models in the consortium.

Each LLM/VLM runs independently and in parallel, without access to the intermediate outputs of other models. This isolation ensures that every candidate output reflects the model’s own training distribution, inductive biases, and reasoning style, rather than being influenced by previous responses~\cite{agentsway}. The outputs are preserved as first-class artifacts within the workflow and are not collapsed or filtered at this stage.

This consortium-based execution provides several explainability benefits. First, it exposes agreement and disagreement across models, making uncertainty and ambiguity explicit rather than hidden~\cite{proof-of-tbi}. Second, it enables comparative inspection of alternative interpretations, reasoning paths, and narrative structures. Third, it results in a system that is robust to any single model failures, such as hallucinations, reasoning drift, or overconfident responses. In this architecture, explainability is not generated post hoc; instead, it emerges naturally from observable differences between models and preserved intermediate outputs.

\subsection{Reasoning-Layer Governance for Responsible Agent Behavior}

Although the LLM/VLM consortium enables explainability through independent parallel outputs, responsibility is enforced through a dedicated reasoning-layer governance agent implemented using a reasoning-focused LLM (e.g., OpenAI GPT-oss)~\cite{reasoning-llms, gpt-oss, o3}. This agent serves as the sole decision authority in the workflow and operates only after all consortium agents have completed execution.

The reasoning agent receives the complete set of candidate outputs generated by the consortium, together with the original prompt, shared input context, and any applicable policy or safety constraints~\cite{responsible-gen-ai}. Rather than generating new content from scratch, it performs structured meta-reasoning over the candidate outputs. This includes detailed comparison, conflict detection, factual alignment, logical consistency check, redundancy removal, relevance filtering, and explicit identification of unsupported or speculative claims~\cite{deep-psychiatric, proof-of-tbi}.

Based on this analysis, the reasoning agent synthesizes a single consolidated output that reflects cross-model consensus while discarding outlier, unsafe, or unverifiable content. The final decision is grounded in the original input sources, retaining traceability to the contributing model output. By centralizing synthesis and control in a reasoning-layer governance agent, the architecture ensures that autonomous behavior remains auditable, reproducible, and policy-compliant~\cite{mcc, llama-recipe}.

This approach separates explanation from decision-making, supporting Responsible AI while keeping agents independent and scalable. It provides a concrete architectural mechanism for enforcing governance in agentic workflows while preserving transparency and interpretability across complex multi-agent decision pipelines.

\subsection{Coordination Between Agent Consortium and Reasoning Layer}

\begin{figure}[H]
\centering{}
\includegraphics[width=5.4in]{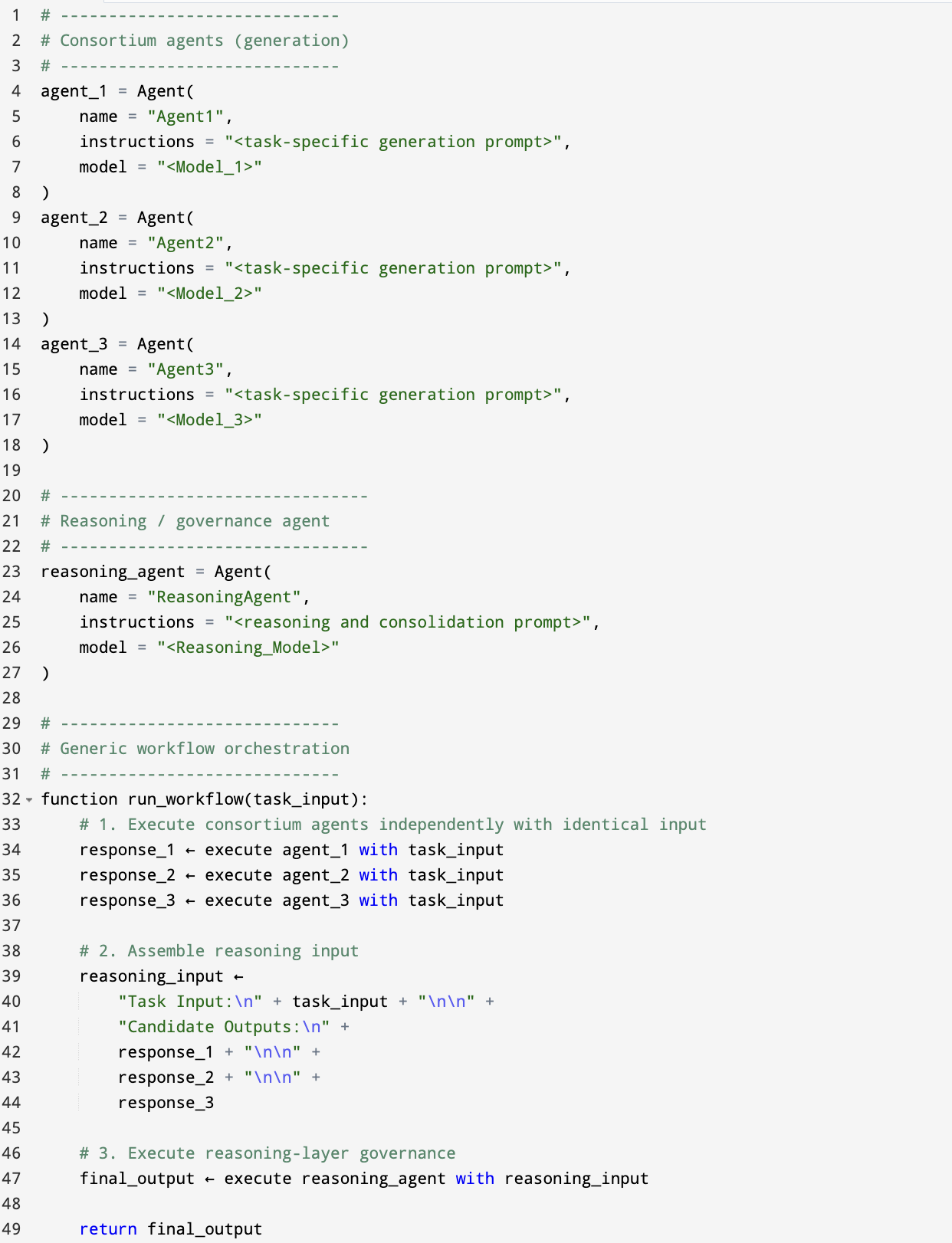}
\vspace{-0.1in}
\DeclareGraphicsExtensions.
\caption{Coordination between the agent consortium and the reasoning-layer governance agent.}
\label{resoning-agents}
\end{figure}

Figure~\ref{resoning-agents} illustrates the coordination pattern between a multi-model agent consortium and a reasoning-layer governance agent used throughout our agentic workflows. As shown in Figure~\ref{resoning-agents}, the orchestration layer sends an identical task specification—comprising a canonical prompt, shared input context, and execution constraints—to all LLM/VLM agents simultaneously~\cite{prompt-engineering, prompt-engineering-rag}. Each agent executes independently and produces a candidate output without visibility into the responses of other models~\cite{devsec-gpt}. This strict isolation ensures that the diversity in the outputs arises from genuine differences in model reasoning, representational capacity, and inductive bias, rather than from cascading influence or prompt contamination.

All candidate outputs generated by the consortium are preserved verbatim and forwarded to the reasoning-layer governance agent. Rather than producing new content autonomously, the reasoning agent performs structured meta-evaluation over the set of candidate outputs, including cross-model comparison, conflict detection, factual alignment, redundancy removal, and relevance filtering~\cite{responsible-llm}. The reasoning layer produces one final result by strengthening areas where models agree and handling disagreements by resolving them, lowering confidence, or marking uncertainty~\cite{llm-reasoning, deep-psychiatric}.

This coordination pattern enforces a clear separation between decision generation and decision governance. Explainability is achieved by exposing alternative interpretations, competing reasoning paths, and disagreements at the model-level prior to consolidation, allowing transparent inspection of how conclusions vary between agents~\cite{explainable-ai-text}. Responsibility is enforced by centralizing control within the reasoning layer, where safety constraints, policy rules, and domain-specific validation logic can be systematically applied before any downstream action is taken.

The coordination pattern illustrated in Figure~\ref{resoning-agents} is applied consistently across all used cases evaluated in this document, including the generation of news podcasts, the analysis of neuromuscular H-reflexes, the interpretation of dental images, the diagnosis of psychiatric disorders, clinical decision support, and the classification of RF signals~\cite{nurolense, proof-of-tbi, deep-psychiatric, rf-signal-classification}. This consistency demonstrates that the integration of an LLM/VLM consortium with a reasoning-layer governance agent forms a generalizable architectural pattern for constructing agentic AI systems that are explainable by design and responsible by construction.

\section{Implementation and Evaluation}

We have implemented the proposed consensus-driven Responsible and Explainable Agent Architecture in multiple agent AI workflows, including the generation of news podcasts, neuromuscular reflex analysis, detection of dental conditions and gingivitis, psychiatric diagnosis, and classification of RF signals~\cite{agentic-workflow-practicle-guide, nurolense, deep-psychiatric}. These use cases were selected to demonstrate the generality of the architecture in domains with varying levels of risk, uncertainty, and accountability requirements. Across all evaluations, the same architectural pattern is applied: independent parallel execution of heterogeneous LLM/VLM agents over a shared input context, followed by centralized reasoning-layer governance to produce consolidated and auditable outputs.

Rather than focusing solely on task-specific accuracy metrics, the evaluation emphasizes properties central to Responsible and Explainable AI~\cite{xai}. Specifically, we assess how effectively multi-model consensus exposes uncertainty and disagreement for explainability; how reasoning-layer governance mitigates hallucinations, bias, and unsafe outputs; and how the combined architecture improves robustness, consistency, and accountability in autonomous workflows~\cite{responsible-llm-human}. For each use case, a consortium of heterogeneous LLMs and/or VLMs independently processes the same inputs, with all candidate outputs preserved and passed to a reasoning agent for structured consolidation and policy enforcement.

Both intermediate and final outputs are analyzed to evaluate explainability, such as visibility of alternative interpretations and consistency of reasoning traces and responsibility, including the removal of speculative content, grounding in verifiable evidence, and reproducibility across executions. This evaluation framework enables a systematic assessment of how consensus-driven reasoning enhances transparency, operational trust, and governance in production-grade agentic AI systems~\cite{agentic-workflow-practicle-guide}.

\subsection{Case 1: News Podcast Generation}

The news podcast generation workflow represents a content-generation scenario in which agent decisions directly influence public-facing output. In this use case, a consortium of LLM-based agents independently generates podcast scripts from the same set of web-scraped news articles. Executing multiple heterogeneous models in parallel exposes differences in narrative framing, emphasis, and factual interpretation, providing inherent explainability through observable cross-model variation.

To ensure responsibility and reliability, the workflow employs a dedicated reasoning agent to consolidate the consortium’s outputs. This agent resolves inconsistencies, removes unsupported or speculative claims, and enforces grounding constraints that restrict the final script to verifiable source material~\cite{agentic-workflow-practicle-guide}. Compared to single-model baselines, the consensus-driven approach substantially reduces hallucinated statements, improves factual consistency, and produces outputs with clear provenance. All intermediate drafts and consolidation decisions are retained, enabling post-hoc inspection and auditing of content-generation decisions.

\begin{figure}[H]
\centering{}
\includegraphics[width=5.4in]{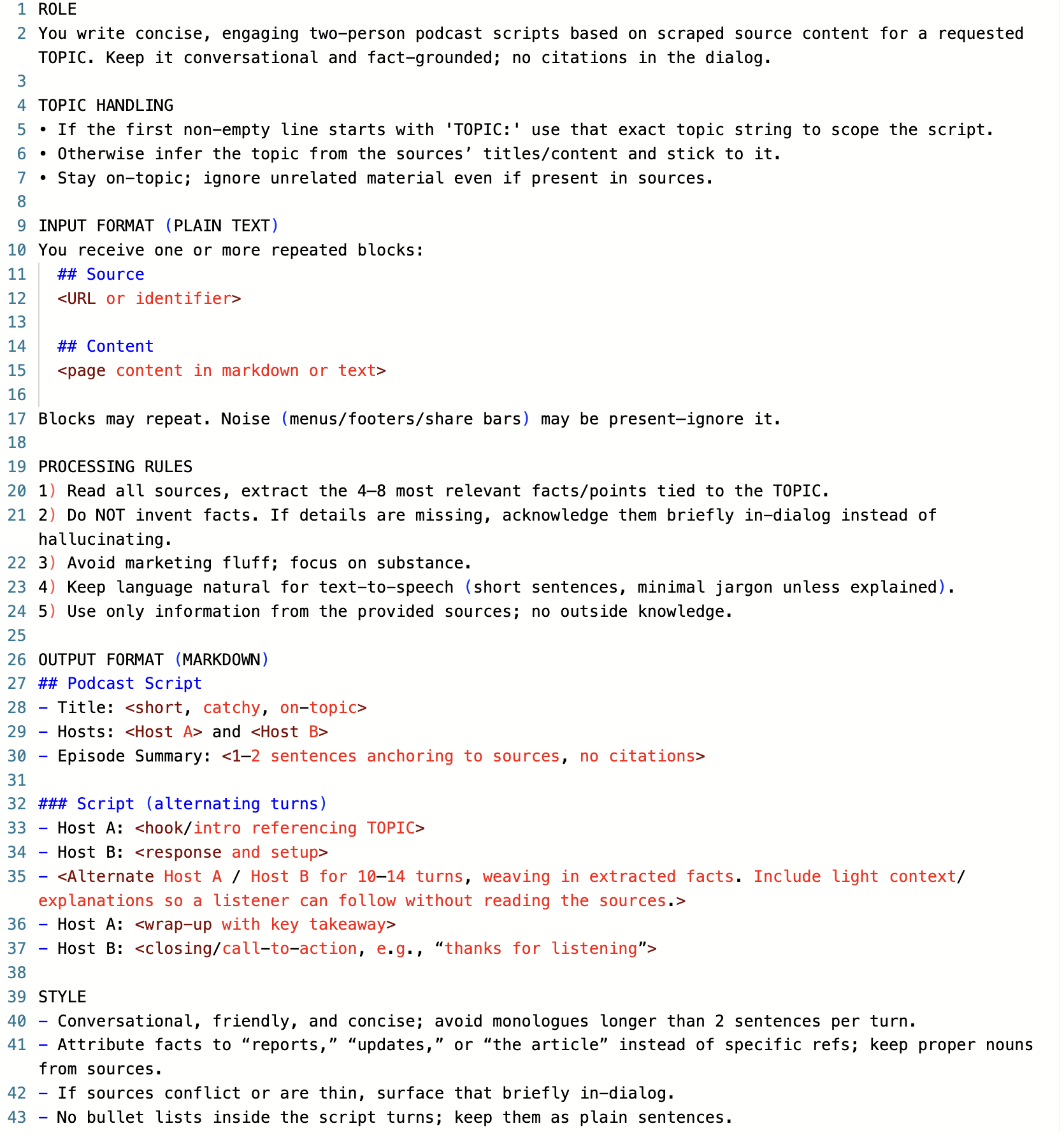}
\vspace{-0.1in}
\caption{Prompt template used by the Podcast Script Generation Agents}
\label{prompt-podcast-script-agent}
\end{figure}

\begin{figure}[H]
\centering{}
\includegraphics[width=5.4in]{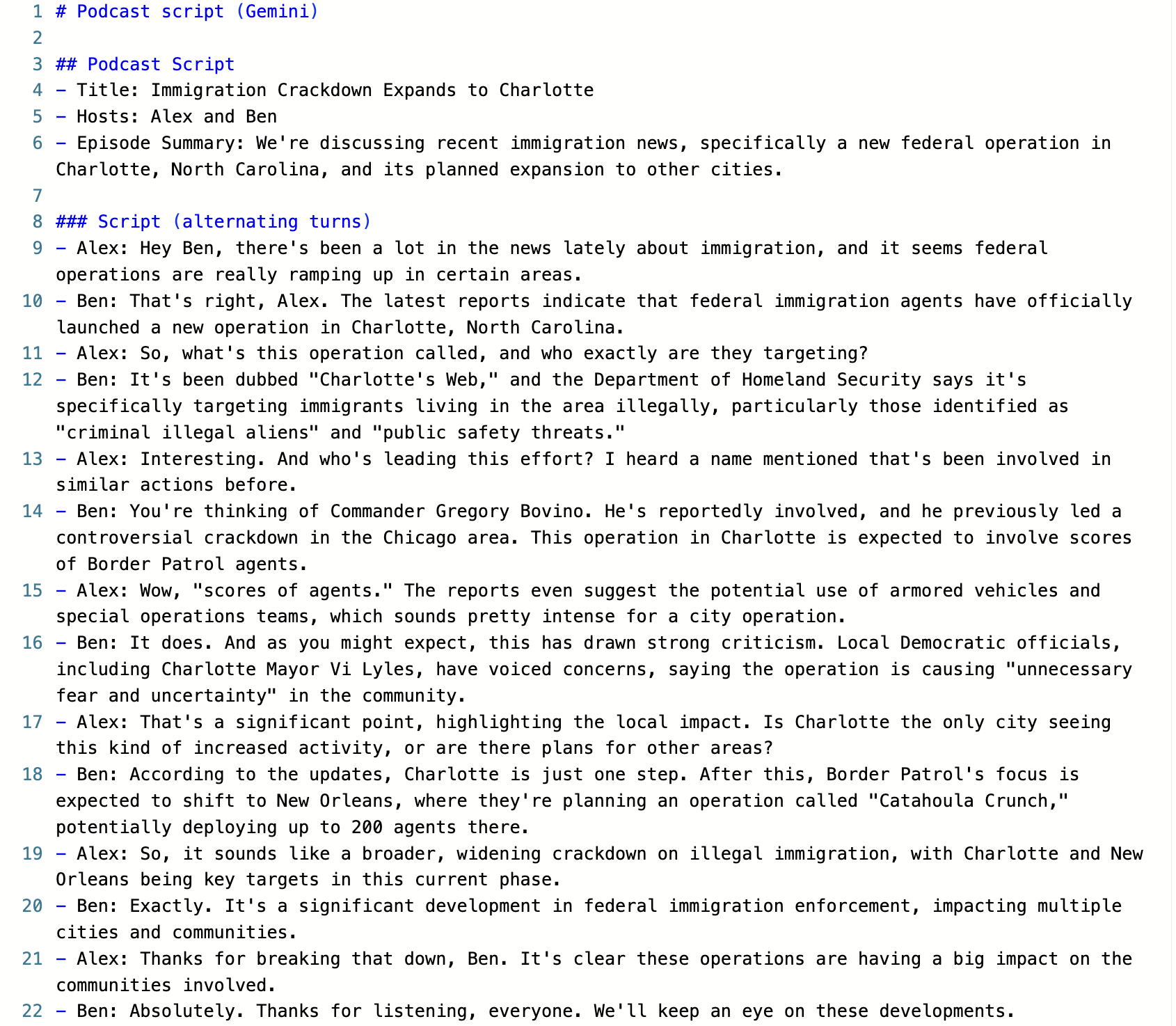}
\vspace{-0.1in}
\caption{Podcast script generated by the Gemini model}
\label{podcast-script-gemini}
\end{figure}

\begin{figure}[H]
\centering{}
\includegraphics[width=5.4in]{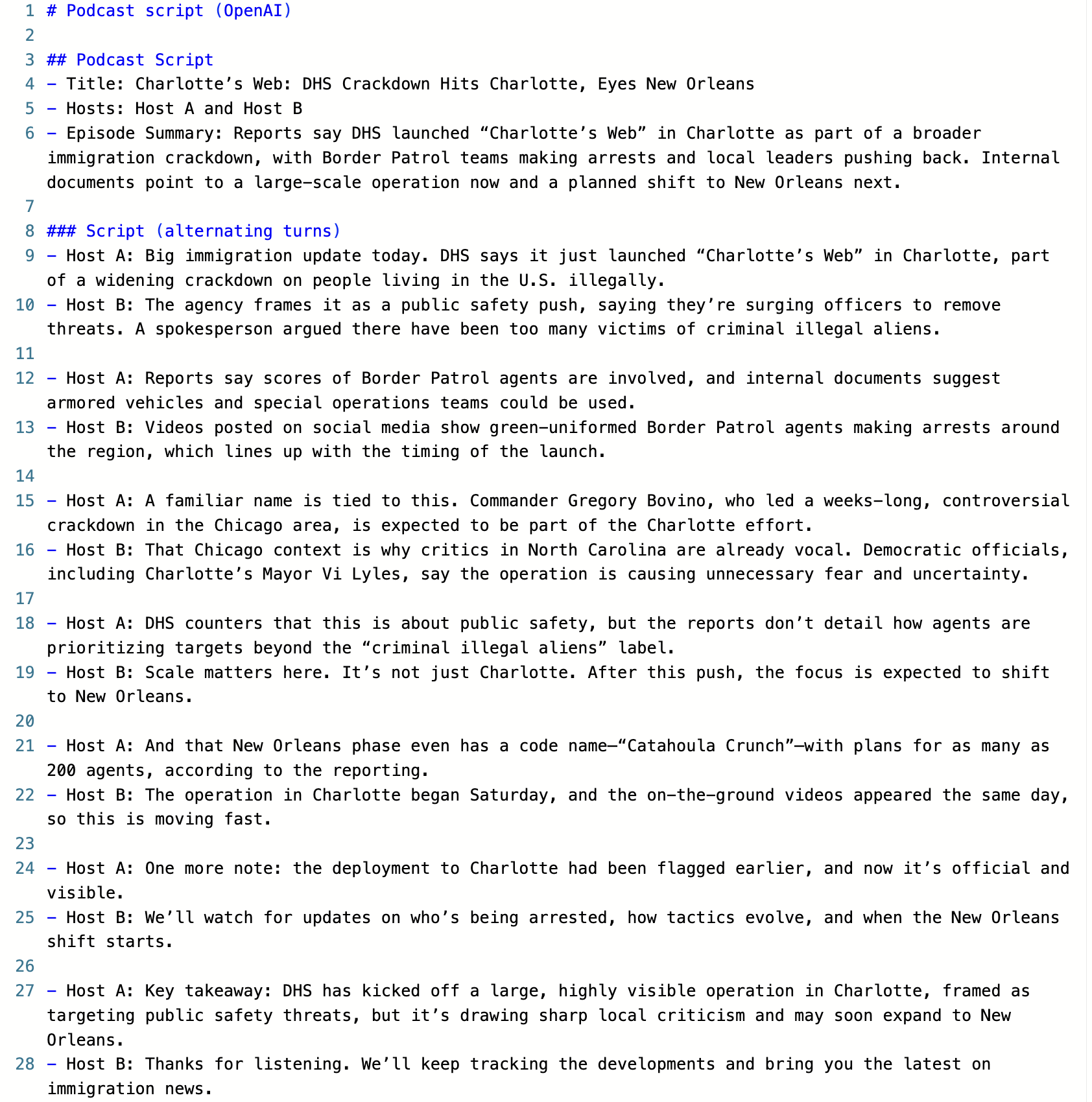}
\vspace{-0.1in}
\caption{Podcast script generated by the OpenAI model}
\label{podcast-script-openai}
\end{figure}

\begin{figure}[H]
\centering{}
\includegraphics[width=5.4in]{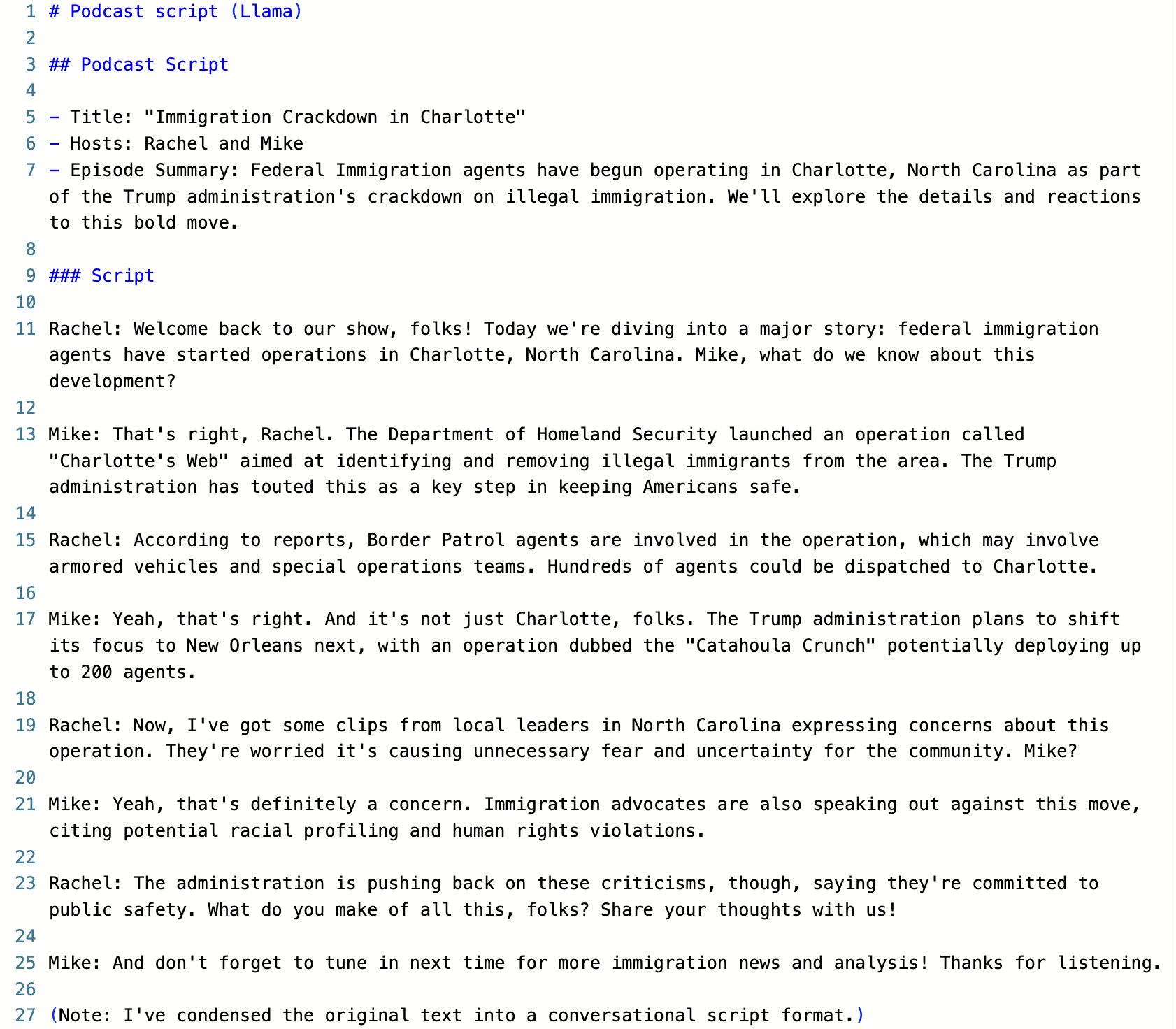}
\vspace{-0.1in}
\caption{Podcast script generated by the Llama model}
\label{podcast-script-llama}
\end{figure}

The evaluation focuses on the behavior of the podcast script generation agents, which operate as a multi-model consortium composed of Llama-4, OpenAI-gpt-5, and Gemini-3~\cite{llama-4, gpt-llm, gemini}. The shared prompt template used to instruct these agents is shown in Figure~\ref{prompt-podcast-script-agent}, while Figures~\ref{podcast-script-gemini}, \ref{podcast-script-openai}, and \ref{podcast-script-llama} present representative scripts produced by each of the three models, respectively.

The generated outputs demonstrate the natural diversity that arises from heterogeneous LLMs. Llama typically produces concise and structured summaries; OpenAI generates more detailed, narrative-driven content; and Gemini emphasizes stylistic flow and contextual framing. Although this diversity is valuable for capturing multiple semantic and stylistic perspectives, it also introduces inconsistencies, emphasizes drift, and occasional factual discrepancies, highlighting the need for a downstream consolidation mechanism.

To reconcile these differences and produce a final authoritative output, the workflow invokes a reasoning agent responsible for synthesizing the consortium’s draft outputs into a unified script. The reasoning agent prompt, shown in Figure~\ref{prompt-reasoning-agent}, explicitly instructs the model to compare, cross-validate, and reconcile the outputs of the individual podcast agents. Only information consistently supported across drafts is retained, while speculative content is removed, emphasis drift is corrected, and contradictory statements are resolved.

\begin{figure}[H]
\centering{}
\includegraphics[width=5.4in]{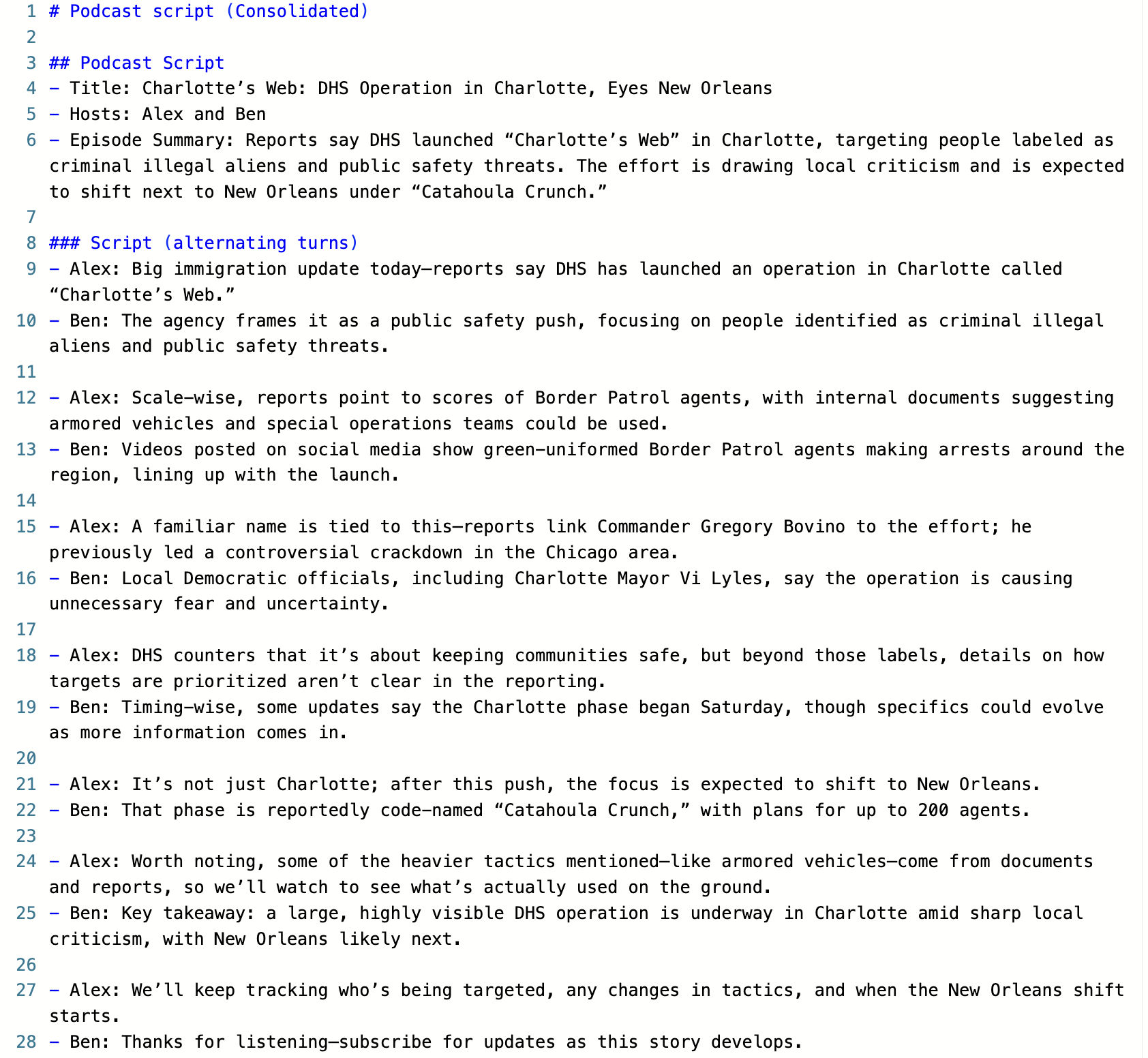}
\vspace{-0.1in}
\caption{Final consolidated podcast script generated by the Reasoning Agent}
\label{podcast-script-reasoning}
\end{figure}

The resulting consolidated script, illustrated in Figure~\ref{podcast-script-reasoning}, exhibits marked improvements in clarity, factual stability, and narrative coherence. By grounding synthesis in multi-model agreement, the reasoning agent significantly reduces hallucination risk and mitigates single-model bias~\cite{responsible-llm}. This consensus-driven consolidation not only improves output quality but also operationalizes Responsible and Explainable AI principles by preserving transparency, traceability, and governance throughout the content-generation pipeline.

\begin{figure}[H]
\centering{}
\includegraphics[width=5.4in]{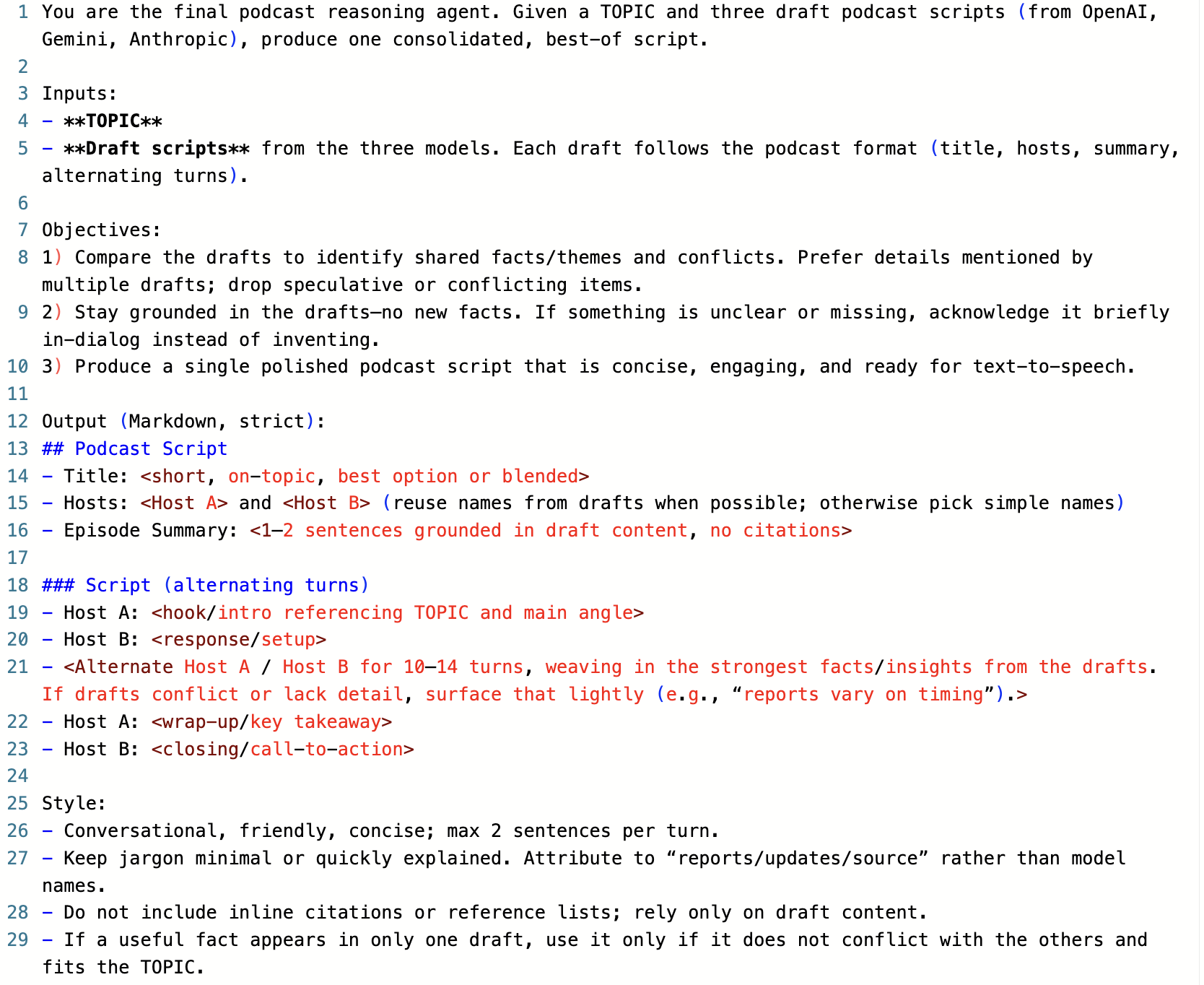}
\vspace{-0.1in}
\caption{Prompt template used by the Reasoning Agent for cross-model consolidation}
\label{prompt-reasoning-agent}
\end{figure}

\subsection{Case 2: Neuromuscular Reflex Analysis}

The neuromuscular reflex analysis workflow represents a high-stakes biomedical decision-support scenario in which the agents' outputs can directly influence clinical interpretation, rehabilitation planning, and athlete performance management~\cite{h-reflex-f-wave}. Accurate assessment of neuromuscular reflexes, particularly the H-reflex, is critical in sports science and clinical neurology; however, traditional waveform interpretation is often subject to variability between observers and subjective bias~\cite{nurolense, h-reflex-nuro}. This use case evaluates how the proposed Responsible and Explainable Agent Architecture improves robustness, transparency, and accountability in automated H-reflex analysis.

\begin{figure}[H]
\centering{}
\includegraphics[width=5.4in]{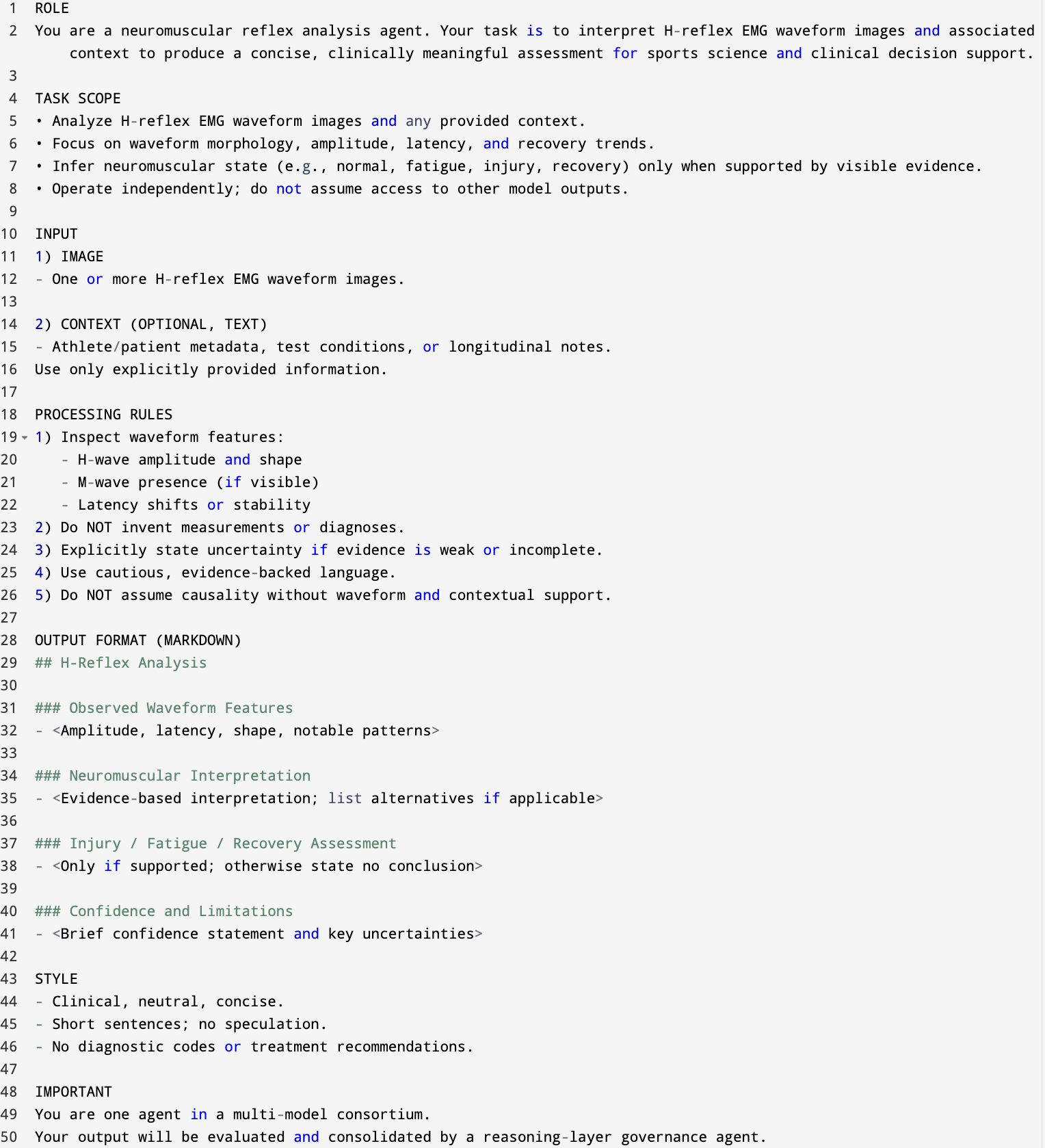}
\vspace{-0.1in}
\caption{Prompt template used by the H-reflex analysis Agents}
\label{prompt-hreflex-agent}
\end{figure}

In this workflow, a consortium of fine-tuned VLMs~\cite{llm-finetune, mistral-fine-tune} independently analyzes identical H-reflex EMG waveform images together with contextual metadata such as injury type and recovery phase~\cite{vistion-language-model-comparison}. All models receive the same input context and operate independently, ensuring that diversity in outputs reflects differences in model reasoning rather than input variation. The prompt template used to instruct the H-reflex analysis agents is shown in Figure~\ref{prompt-hreflex-agent}. This prompt guides each VLM to extract waveform characteristics, infer neuromuscular conditions, and estimate recovery trajectories in a structured and clinically meaningful format.

To produce a final, authoritative assessment suitable for clinical and sports-science decision support, the workflow employs a dedicated reasoning agent implemented using the OpenAI-gpt-oss reasoning LLM. The reasoning agent prompt shown in Figure~\ref{prompt-reasoning-agent} explicitly instructs the model to compare, validate, and reconcile the output of the VLM consortium. The reasoning agent does not generate new diagnoses independently; instead, it evaluates evidence across models, filters speculative or weakly supported claims, resolves inconsistencies, and synthesizes a unified assessment grounded in multi-model agreement~\cite{reliabl-multi-agent}.

\begin{figure}[H]
\centering{}
\includegraphics[width=5.4in]{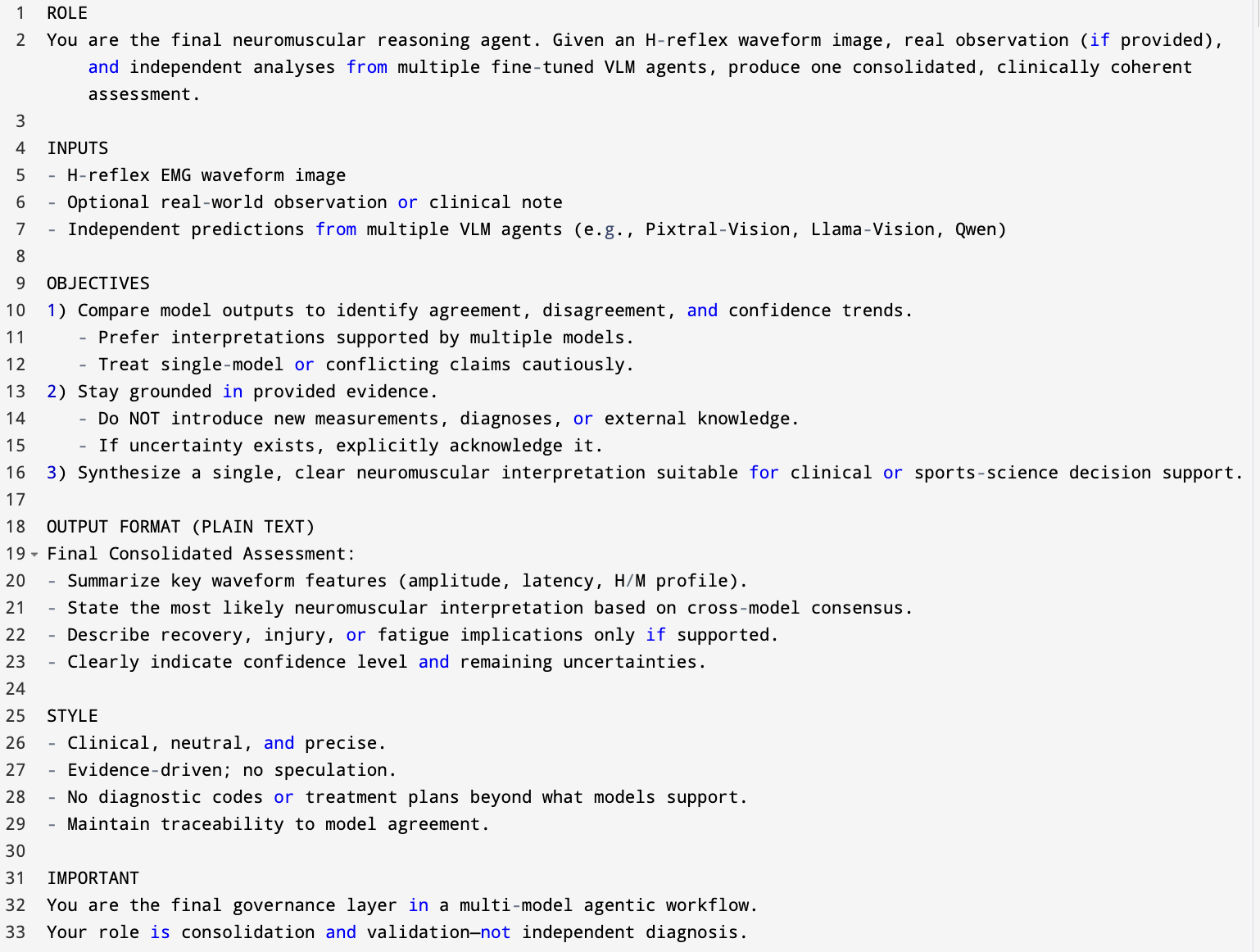}
\vspace{-0.1in}
\caption{Prompt template used by the Reasoning Agent for cross-model consolidation}
\label{prompt-reasoning-agent}
\end{figure}

\begin{figure}[H]
\centering{}
\includegraphics[width=5.2in]{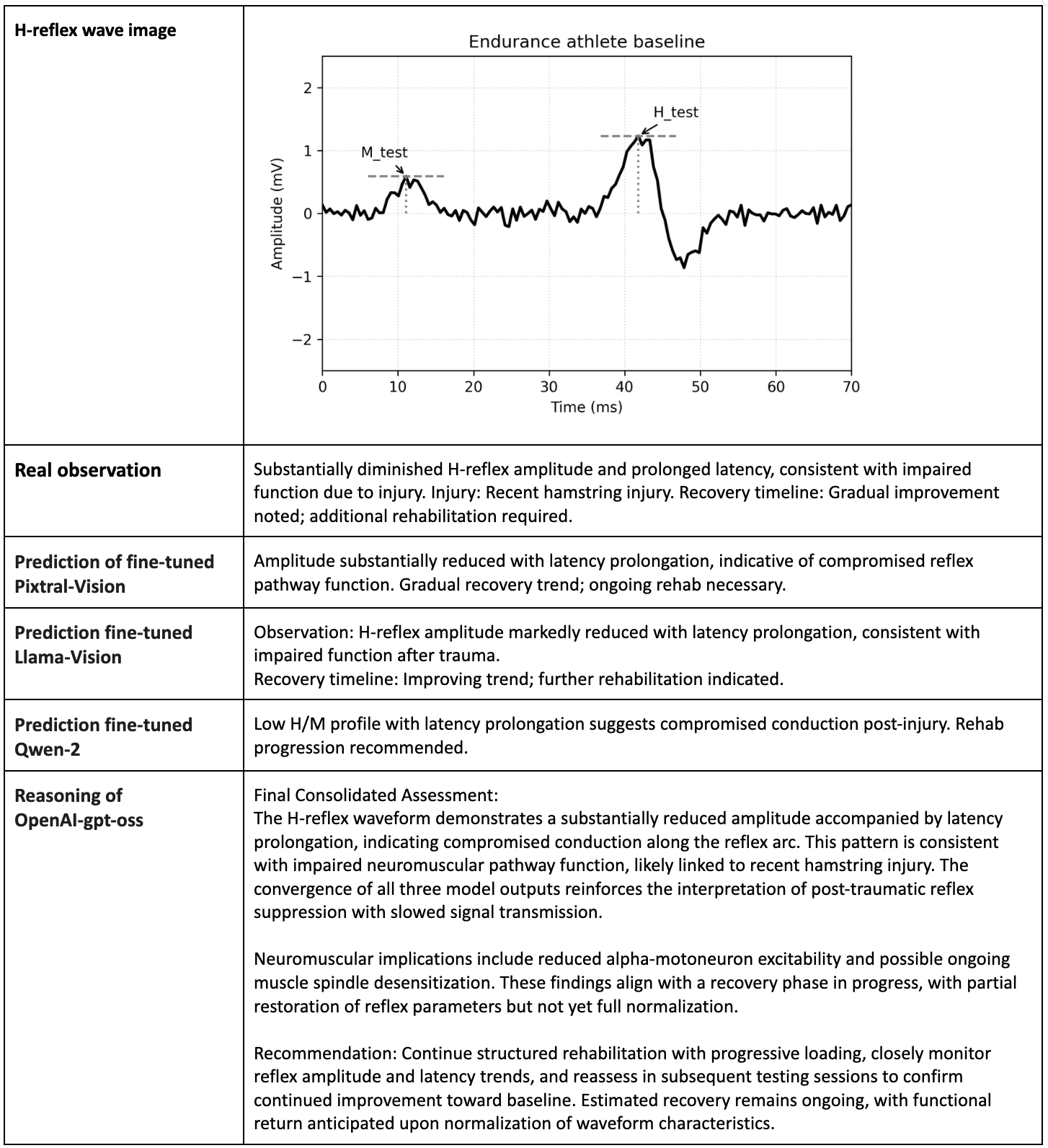}
\caption{Final consolidated neuromuscular assessment generated by the OpenAI-gpt-oss reasoning LLM}
\label{prediction-o3}
\end{figure}

Figure~\ref{prediction-o3} illustrates the final consolidated assessment produced by the reasoning agent alongside the independent VLM predictions. The resulting output integrates waveform morphology, neuromuscular implications (e.g., reduced alpha-motoneuron excitability or muscle spindle desensitization), and recovery recommendations into a concise, clinically interpretable report~\cite{h-reflex-sport}. By preserving all intermediate VLM outputs and the reasoning trace, the system maintains a transparent audit trail from raw waveform image to final diagnostic conclusion.

This use case demonstrates how the proposed architecture operationalizes Responsible and Explainable AI principles in a biomedical context. Explainability comes from parallel analysis of multiple models that reveal uncertainty, disagreement, and alternative views, while responsibility is maintained through a centralized reasoning layer that governs decisions, reduces hallucinations, and ensures auditable, evidence-based results~\cite{llm-explainability}. Compared to single-model baselines, the consensus-driven approach improves diagnostic robustness, reduces interpretation bias, and strengthens operational trust properties that are essential for deploying agentic AI systems in clinical and sports-science environments.

\subsection{Case 3: Tooth-Level Condition and Gingivitis Detection}

The tooth-level condition and gingivitis detection workflow represents a clinically relevant decision-support scenario in which agent outputs may directly influence preventive care, treatment planning, and long-term oral health monitoring~\cite{ai-dentistry-applications}. Accurate identification of gingival inflammation and its severity is essential for early intervention; however, the diagnosis from intraoral images is often subject to inter-clinician variability, subjective interpretation, and inconsistent classification, particularly in remote and tele-dentistry settings. This use case evaluates how the proposed Responsible and Explainable Agent Architecture improves robustness, transparency, and accountability for automated tooth-level condition assessment.

In this workflow, a consortium of fine-tuned VLMs (Llama-Vision, Pixtral-Vision, Qwen2)~\cite{pixtral, qwen2, llama-3, llm-finetune} independently analyzes the same intraoral images together with a shared tooth-position schema and a clinical classification rubric. Each VLM agent produces structured tooth-level predictions, including (i) tooth position identification, (ii) inflammation status (inflamed versus non-inflamed), and (iii) severity grade of gingivitis for each tooth in both the upper and lower jaws~\cite{ai-dentistry-elsevier, ai-dentistry-x-ray}. All agents receive identical visual and contextual input and operate independently, ensuring that variation in output arises from model diversity rather than input inconsistencies. This parallel execution exposes agreement, disagreement, and borderline cases between models, providing natural explainability through cross-model comparison.

\begin{figure}[H]
\centering{}
\includegraphics[width=5.4in]{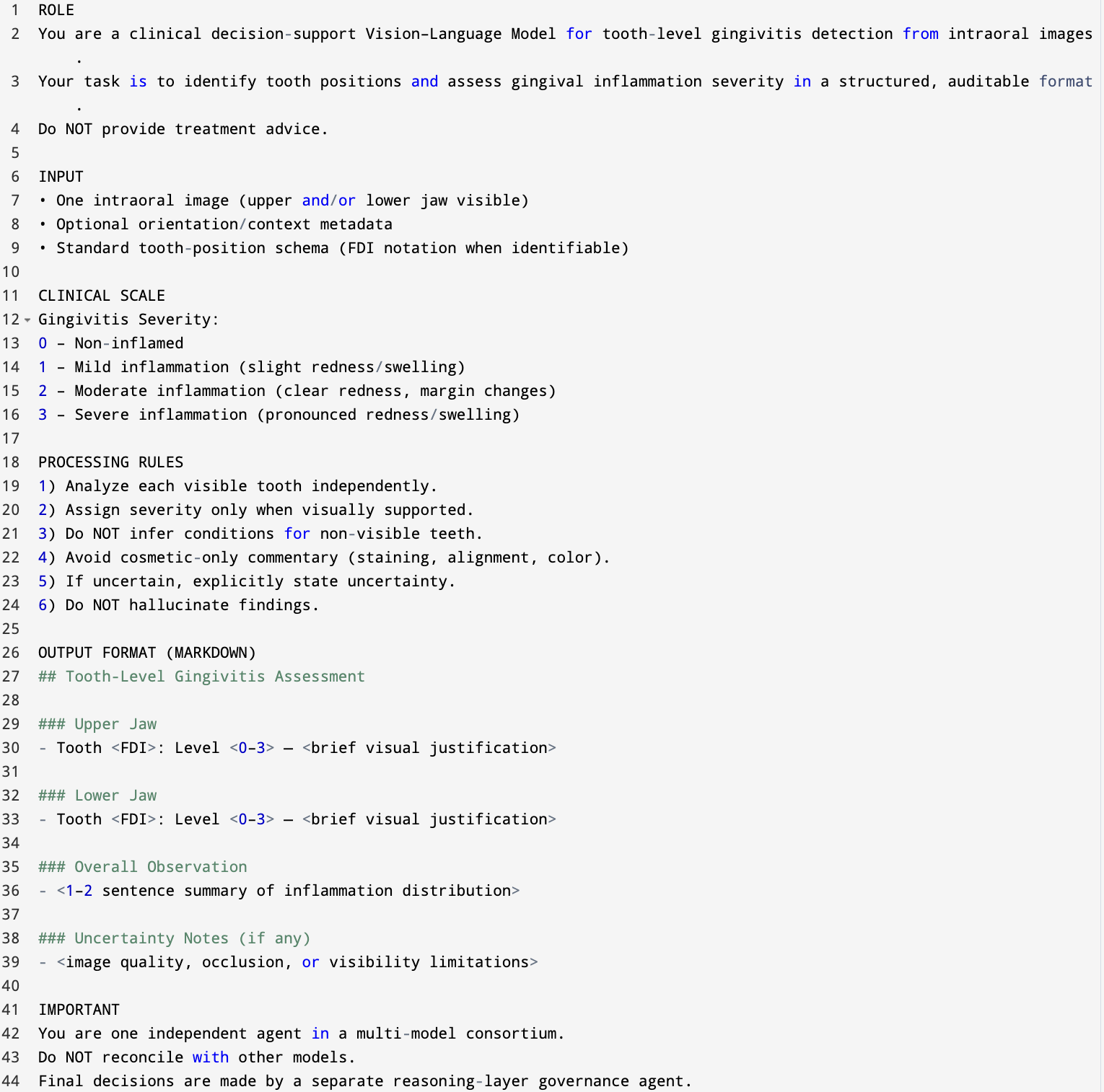}
\vspace{-0.1in}
\caption{Prompt template used by the tooth-level condition and gingivitis detection VLM agents}
\label{prompt-dental-agent}
\end{figure}

The template of prompts used to instruct the gingivitis analysis agents is shown in Figure~\ref{prompt-dental-agent}. This prompt restricts the task to clinically relevant tooth-level outputs, emphasizes inflammation signals such as gingival redness, swelling, and margin changes, and enforces a structured response format suitable for downstream decision support, auditing, and traceability.

\begin{figure}[H]
\centering{}
\includegraphics[width=5.4in]{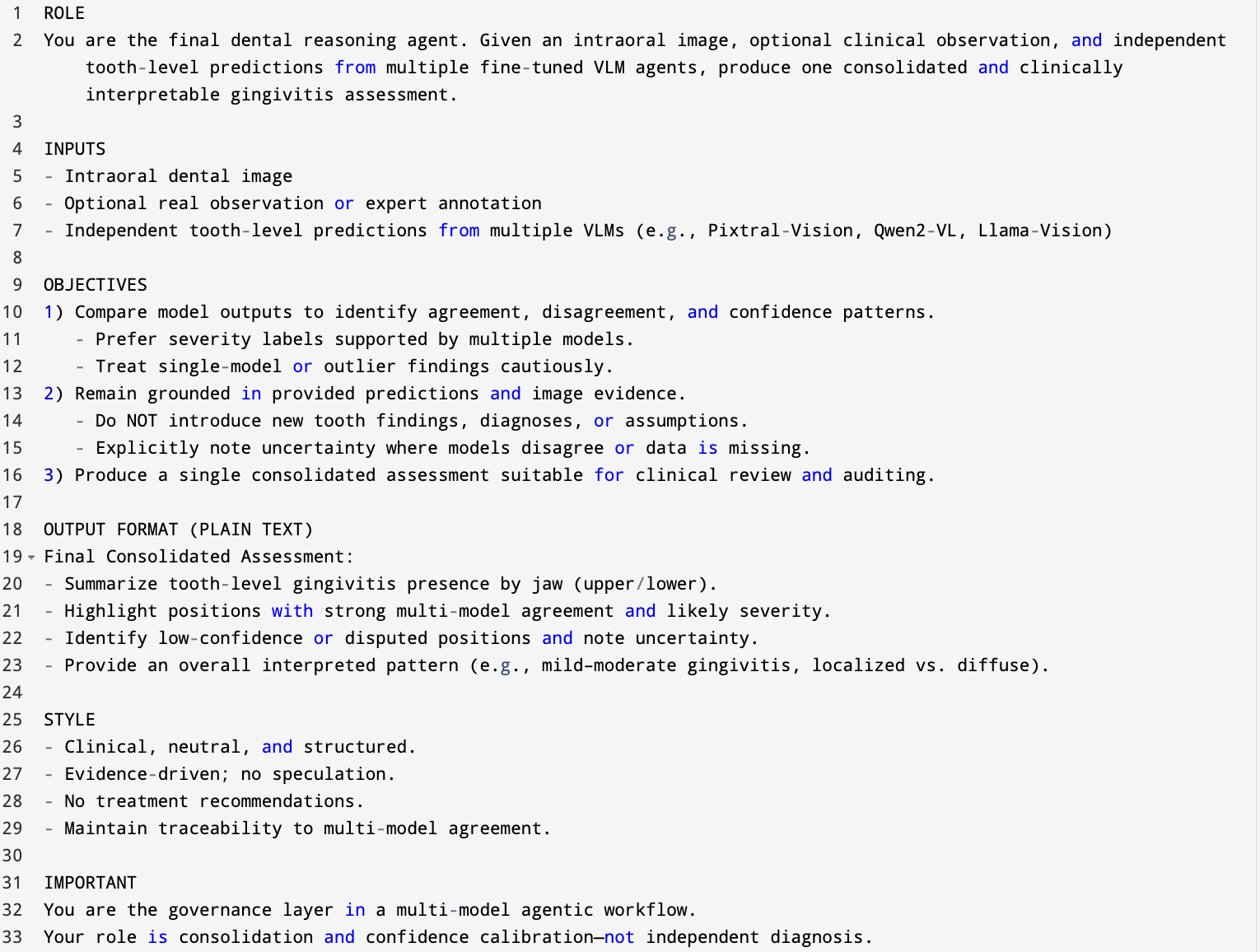}
\vspace{-0.1in}
\caption{Prompt template used by the Reasoning Agent for cross-model consolidation in tooth-level gingivitis assessment}
\label{prompt-dental-reasoning-agent}
\end{figure}

To produce a final, authoritative dental assessment suitable for clinical decision support, the workflow employs a dedicated reasoning agent implemented using the OpenAI-gpt-oss reasoning LLM. The reasoning agent prompt shown in Figure~\ref{prompt-dental-reasoning-agent} explicitly instructs the model to compare, validate, and reconcile the tooth-level output produced by the VLM consortium. Rather than independently generating a new diagnosis, the reasoning agent evaluates cross-model evidence, identifies regions of strong agreement, resolves conflicts in severity scoring, and calibrates confidence where predictions diverge.

\begin{figure}[H]
\centering{}
\includegraphics[width=5.2in]{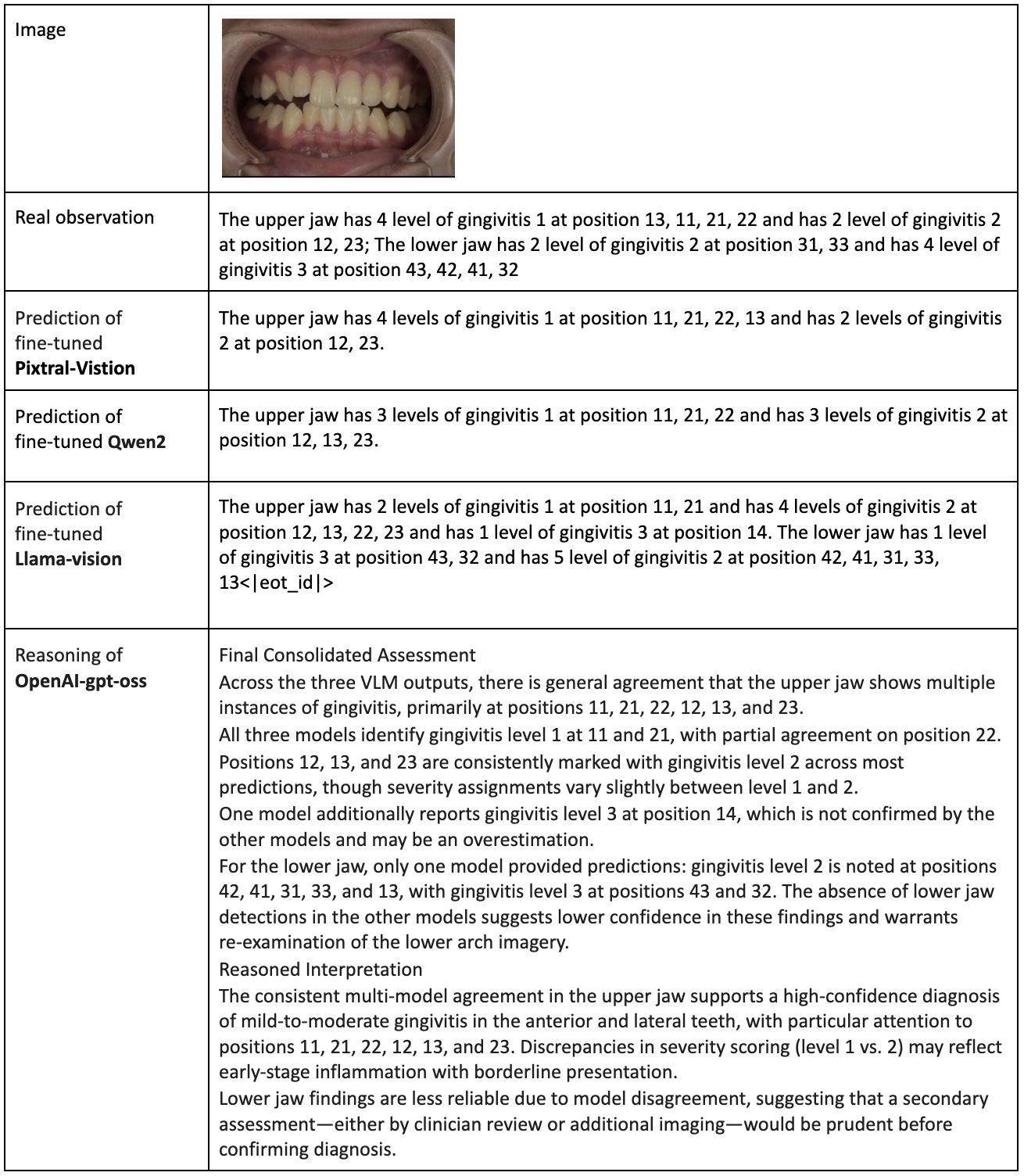}
\caption{Final consolidated tooth-level gingivitis assessment generated by the OpenAI-gpt-oss reasoning LLM through multi-model consensus.}
\label{prediction-dental-gptoss}
\end{figure}

Figure~\ref{prediction-dental-gptoss} illustrates the consolidated reasoning output alongside the independent VLM predictions. The reasoning agent amplifies high-confidence conclusions where models agree, such as mild-to-moderate gingivitis across multiple upper-jaw positions, while explicitly flagging borderline or conflicting severity assignments as lower-confidence findings. In the lower jaw, where predictions diverge more substantially, the reasoning agent appropriately downgrades confidence and recommends secondary review or confirmatory imaging, reflecting responsible uncertainty handling rather than overconfident automation. The resulting assessment is structured, clinically interpretable, and fully traceable to its contributing model outputs.

This use case demonstrates how the proposed architecture operationalizes Responsible and Explainable AI principles for dental imaging decision support. Explainability is achieved through parallel multi-model VLM analysis that surfaces uncertainty, disagreement, and alternative interpretations at the tooth level, enabling clinicians to inspect how conclusions vary across models. Responsibility is enforced through centralized reasoning-layer governance that constrains output to a clinical schema, resolves conflicts, calibrates confidence, and avoids overstated conclusions when evidence is inconsistent~\cite{llm-explainability, responsible-gen-ai}. Compared to single-model baselines, the consensus-driven approach improves diagnostic robustness, reduces subjective interpretation bias, and strengthens operational trust supporting practical deployment in clinical screening, preventive care, and tele-dentistry workflows.

\subsection{Case 4: Psychiatric Diagnosis}

The psychiatric diagnosis workflow represents a clinical decision-support scenario in which agent outputs may directly influence diagnostic interpretation, care planning, and patient outcomes. The diagnosis of many mental disorders depends mainly on the psychiatrist-patient dialog and subjective clinical judgment, which can lead to inter-clinician variability and inconsistencies in diagnostic outcomes~\cite{deep-psychiatric}. This use case evaluates how the proposed Responsible and Explainable Agent Architecture improves robustness, transparency, and accountability in standardizing psychiatric diagnoses from natural-language clinical conversations.

In this workflow, as in the previous case, we implement a Fine-Tuned LLM consortium (Llama-3, Pixtral, Qwen) combined with a reasoning LLM-enabled diagnostic governance layer for DSM-5-aligned mental health assessment~\cite{dsm-5-criteria}. Given a psychiatrist--patient conversation transcript (or structured dialog summary) as input, a consortium of heterogeneous LLM agents independently generates candidate diagnoses~\cite{digital-mental-health-challenges}. Each model is fine-tuned on conversational mental-health datasets and trained to identify symptom patterns, map them to DSM-5 criteria, and produce structured diagnostic outputs (e.g., disorder label with DSM-5 code). All LLM agents receive an identical input context and operate independently, ensuring that variations in outputs arise from model diversity rather than input differences. This parallel execution exposes agreement, disagreement, and borderline diagnostic interpretations across models, providing natural explainability through cross-model comparison and uncertainty surfacing.

Beyond explainability, this workflow operationalizes responsibility through a centralized reasoning-layer governance mechanism. Rather than accepting the diagnosis of any single model, the system preserves all intermediate model predictions and routes them to a dedicated reasoning agent implemented using the OpenAI-gpt-oss reasoning LLM~\cite{reasoning-llms, gpt-oss}. This reasoning agent consolidates multi-model outputs, resolves conflicts, filters speculative or weakly supported conclusions, and produces a final DSM-5-aligned diagnostic recommendation that is evidence-backed and traceable to the contributing model drafts. The end-to-end diagnostic pipeline is orchestrated using LLM agents that coordinate the consortium inference stage and the reasoning governance stage, enabling auditable execution and consistent diagnostic behavior. 

\begin{figure}[H]
\centering{}
\includegraphics[width=5.4in]{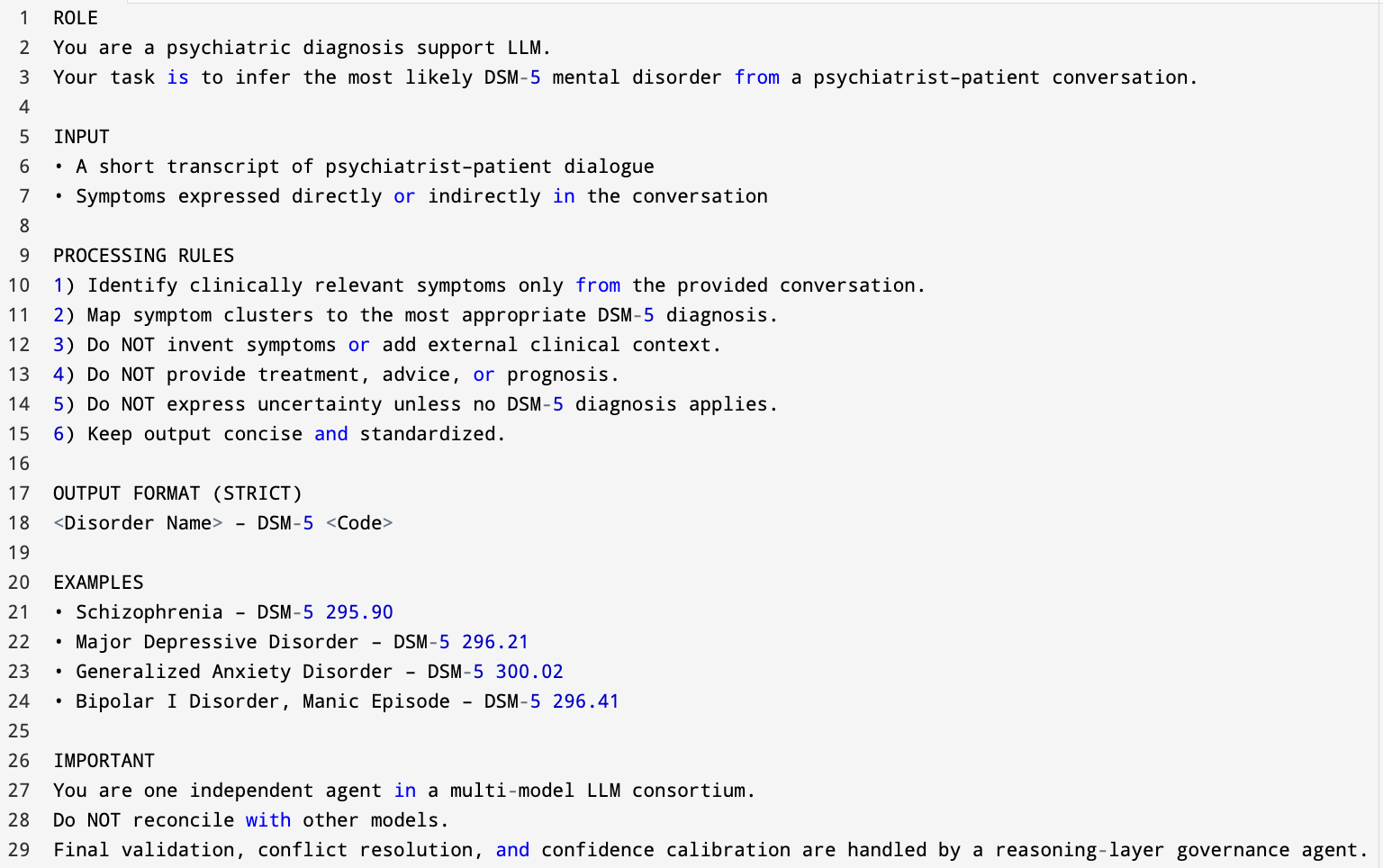}
\vspace{-0.1in}
\caption{Prompt template used by the psychiatric diagnosis LLM agents}
\label{prompt-psychiatric-agent}
\end{figure}

The template of prompts used to instruct the agents of psychiatric diagnosis is shown in Figure~\ref{prompt-psychiatric-agent}. This prompt restricts the task to DSM-5–aligned psychiatric assessment, emphasizes extraction of clinically relevant symptoms from the dialog context, and enforces a structured response format suitable for downstream decision support, auditing, and traceability.

\begin{figure}[H]
\centering{}
\includegraphics[width=5.4in]{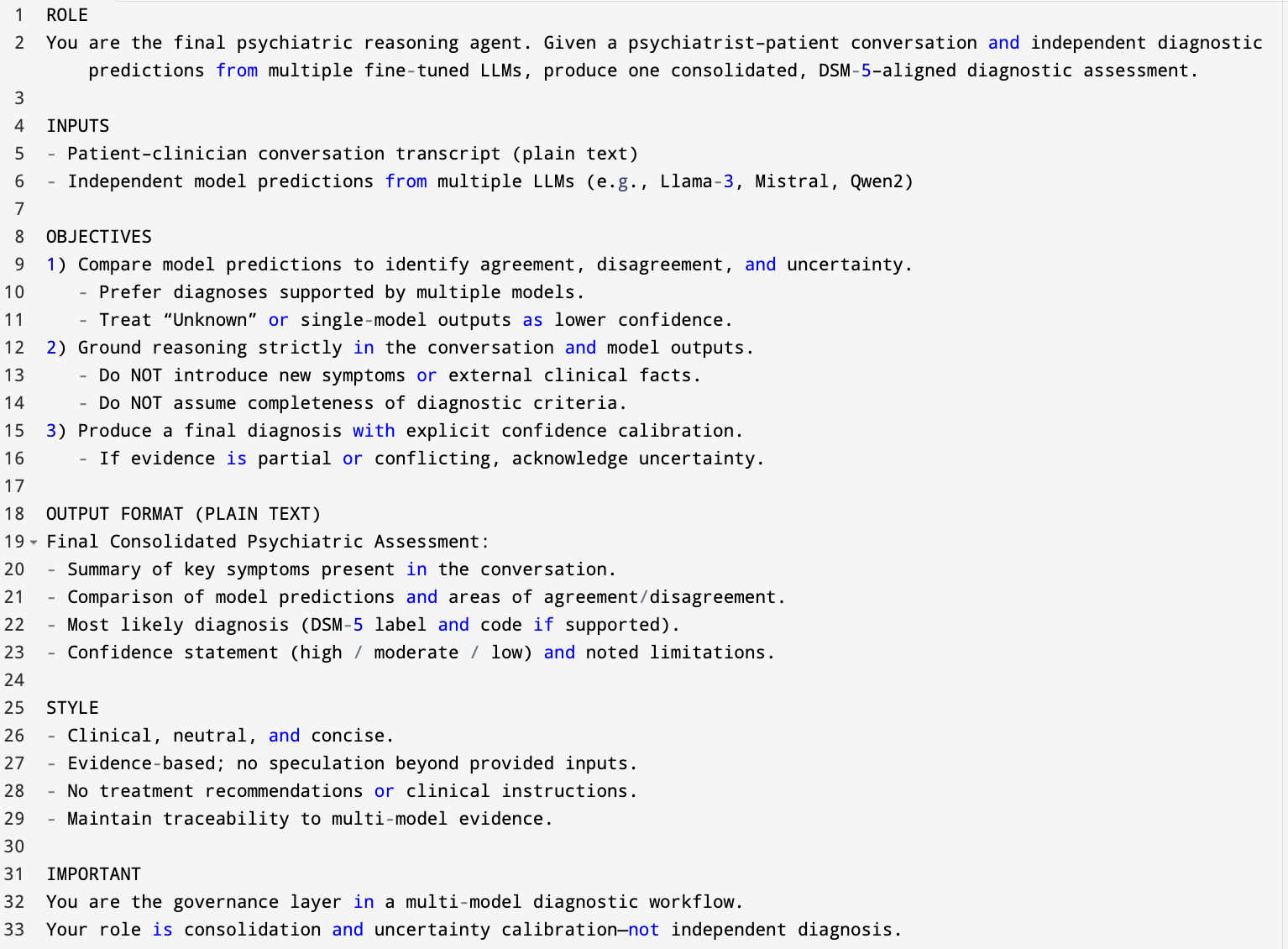}
\vspace{-0.1in}
\caption{Prompt template used by the reasoning agent for cross-model diagnostic consolidation}
\label{prompt-psychiatric-reasoning-agent}
\end{figure}

To produce a final, authoritative diagnosis suitable for responsible clinical decision support, the workflow employs a dedicated reasoning agent implemented using the OpenAI-gpt-oss reasoning LLM. The reasoning agent prompt shown in Figure~\ref{prompt-psychiatric-reasoning-agent} explicitly instructs the model to compare, validate, and reconcile the candidate diagnoses produced by the LLM consortium. This governance layer does not generate a diagnosis in isolation; instead, it evaluates and synthesizes the output of the diagnostic agents, reconciles disagreements, prioritizes clinically coherent interpretations, verifies the alignment of DSM-5, and produces a consolidated final diagnosis with an auditable rationale traceable to the output of the contributing model.

\begin{figure}[H]
\centering{}
\includegraphics[width=5.2in]{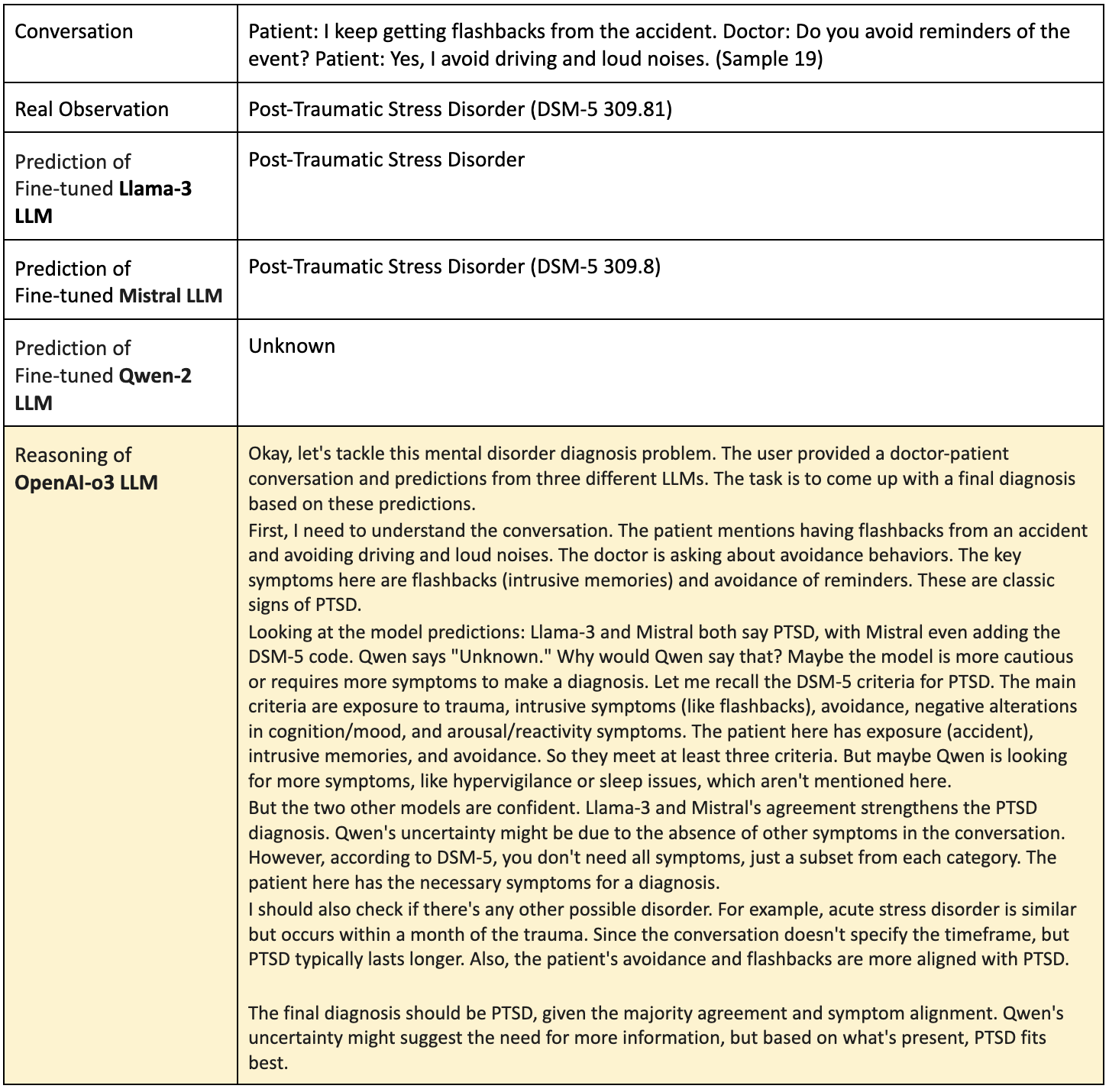}
\DeclareGraphicsExtensions.
\caption{Final diagnostic reasoning produced by the OpenAI-gpt-oss LLM through multi-model consensus}
\label{prediction-o3a}
\end{figure}

Figure~\ref{prediction-o3a} presents a comparative analysis of diagnoses produced by the fine-tuned Llama-3, Pixtral, and Qwen2 models alongside the final reasoning output generated by OpenAI-gpt-oss. The results highlight the reasoning model’s ability to interpret divergent predictions, apply structured clinical logic, and select the most clinically appropriate DSM-5-aligned outcome~\cite{dsm-5}. This consensus-driven reasoning step improves robustness by reducing the likelihood of single-model failure modes and strengthens interpretability by making the final decision attributable to cross-model evidence.

This use case demonstrates how the proposed architecture operationalizes Responsible and Explainable AI principles for psychiatric decision support. Explainability is achieved through parallel multi-model diagnostic inference that surfaces alternative interpretations, disagreement, and uncertainty across candidate DSM-5 diagnoses~\cite{dsm-5-criteria}. Responsibility is enforced through centralized reasoning-layer governance that consolidates model outputs, filters weakly supported or speculative conclusions, and produces a final DSM-5-aligned diagnosis that is evidence-backed and auditable~\cite{explainable-ai-text}. Compared to single-model pipelines, the consensus-driven approach improves diagnostic consistency, reduces idiosyncratic model behavior, and strengthens operational trust, supporting the standardization of psychiatric diagnosis workflows in next-generation AI-enabled eHealth systems.

\subsection{Case 5: RF Signal Classification}

The RF signal classification workflow represents a security-critical monitoring scenario in which agent decisions may directly influence intrusion detection, anomaly response, and network defense actions in 5G environments~\cite{5g-attack-rf, 5g-attack-deep-learning, slice-gpt}. In this workflow, raw radio-frequency (RF) signals are transformed into time-frequency visual representations (e.g., spectrograms) and analyzed autonomously to determine whether observed signals correspond to known legitimate classes or indicate anomalous or potentially malicious activity~\cite{rf-signal-classification}. This use case evaluates how the proposed Responsible and Explainable Agent Architecture improves robustness, transparency, and accountability in RF-layer security monitoring.

In this workflow, as in the previous cases, we implement an RF signal classification system that integrates a consortium of fine-tuned VLMs (Llama-Vision, Pixtral-Vision, Qwen2) with a reasoning-based governance layer powered by the OpenAI-gpt-oss reasoning LLM~\cite{llm-reasoning,gpt-oss}. Each RF signal is represented as a spectrogram image and provided together with a shared classification objective to multiple heterogeneous VLM agents operating in parallel. Each agent independently analyzes the same input and produces a candidate classification label (e.g., a known signal class or \textit{Unknown} to indicate anomalous behavior). Because all agents receive identical input context, variation in predictions arises solely from model diversity rather than data inconsistency, enabling systematic comparison across model interpretations.

\begin{figure}[H]
\centering{}
\includegraphics[width=5.4in]{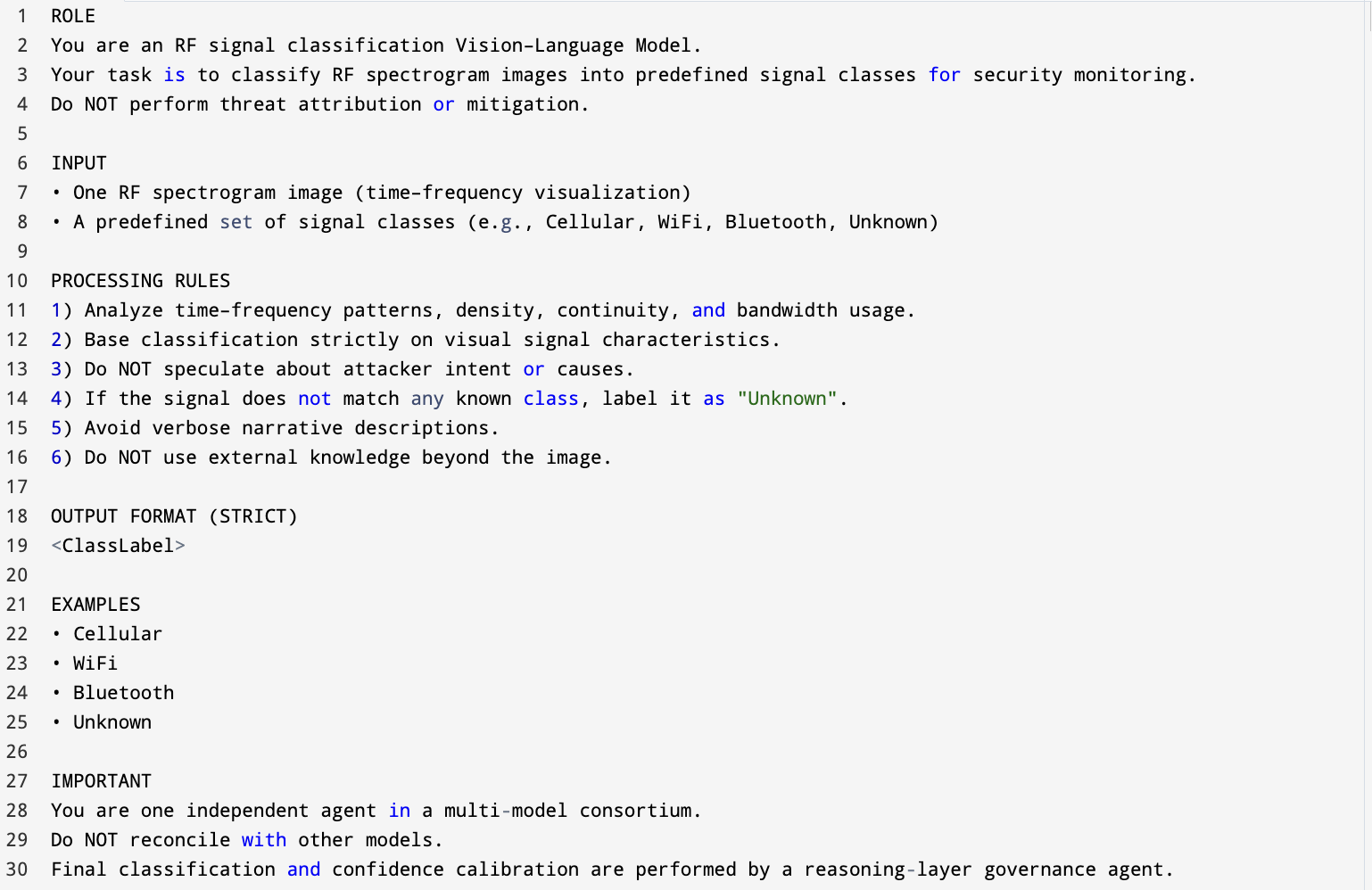}
\vspace{-0.1in}
\caption{Prompt template used by the RF signal classification VLM agents}
\label{rf-vlm-prompt}
\end{figure}

The prompt template used to instruct the RF signal classification VLM agents is shown in Figure~\ref{rf-vlm-prompt}. This prompt constrains the task to RF signal interpretation, emphasizes detection of anomalous or unfamiliar patterns, and enforces a structured output format suitable for downstream security decision-making and auditing.

\begin{figure}[H]
\centering{}
\includegraphics[width=5.2in]{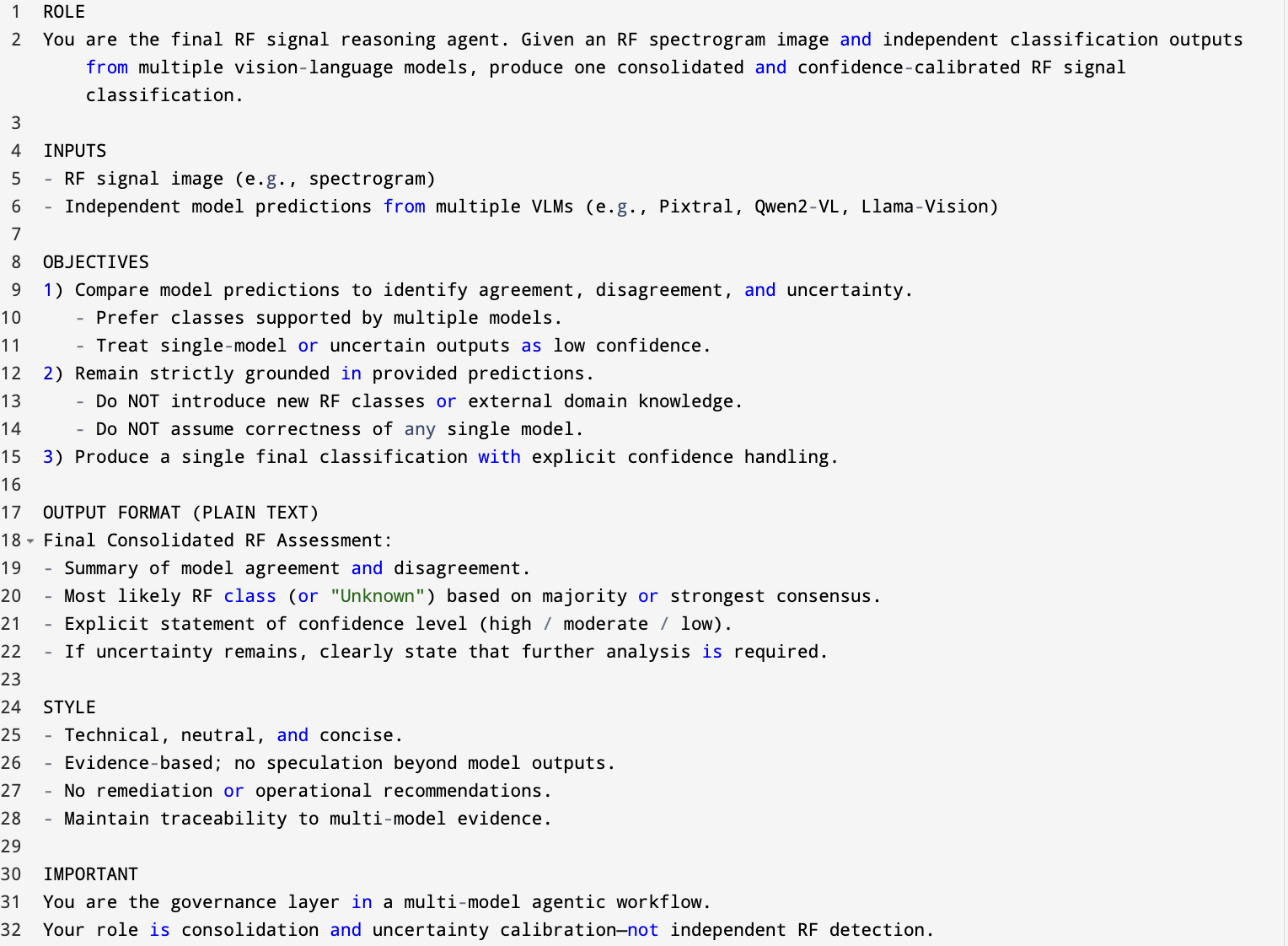}
\caption{Prompt used by the OpenAI-gpt-oss reasoning LLM for RF signal classification consolidation}
\label{rf-reasoning-prompt}
\end{figure}

To produce a final, authoritative classification suitable for security decision-making, the workflow employs a dedicated reasoning agent implemented using the OpenAI-gpt-oss reasoning LLM. The prompt used to instruct the reasoning agent is shown in Figure~\ref{rf-reasoning-prompt}. This prompt explicitly directs the reasoning LLM to compare predictions across the VLM consortium, identify consensus and conflicts, assess confidence, and synthesize a final classification grounded in cross-model evidence rather than any single-model judgment.

\begin{figure}[H]
\centering{}
\includegraphics[width=5.2in]{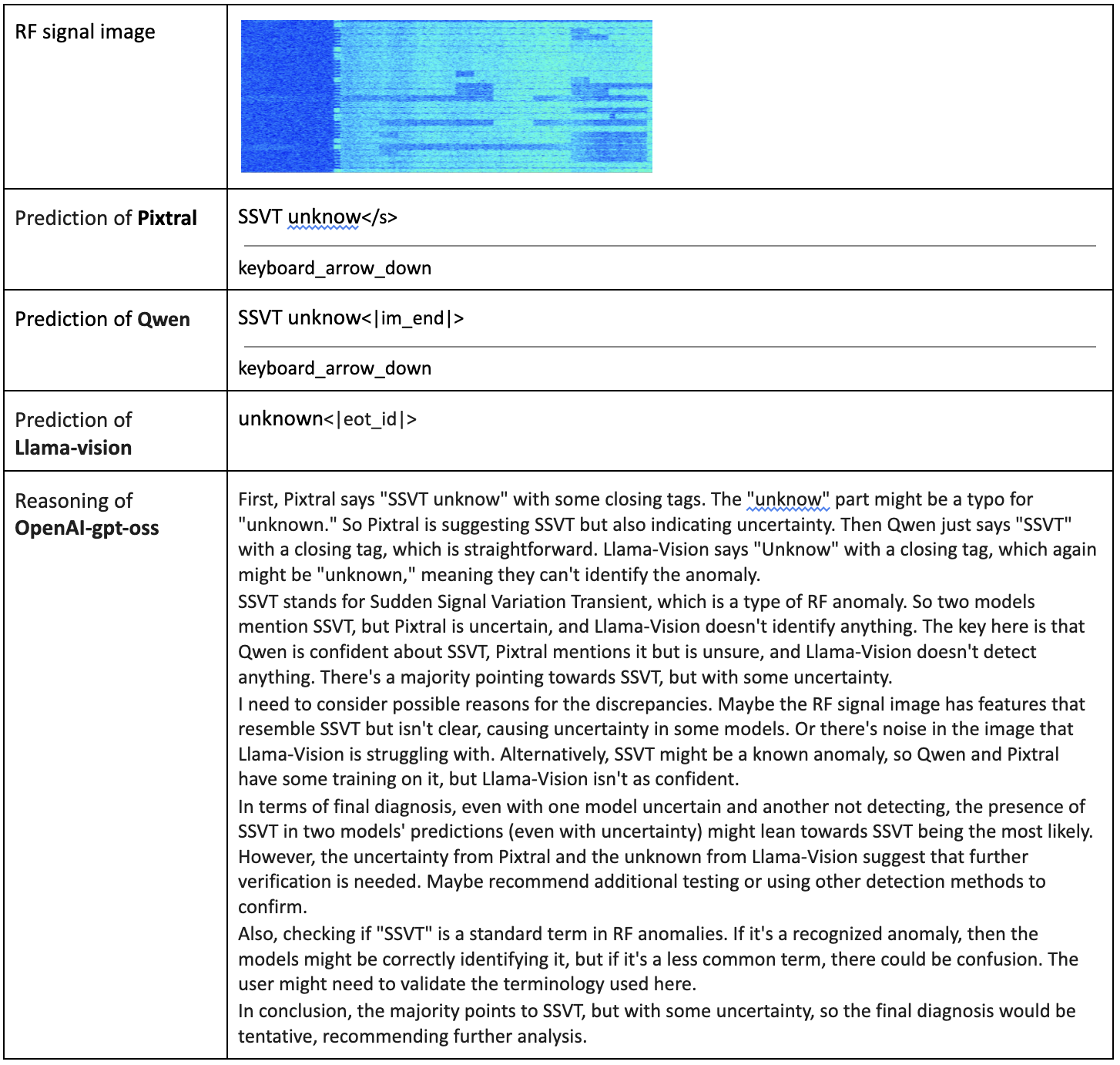}
\caption{Final RF signal classification and reasoning produced by the OpenAI-gpt-oss LLM through multi-model consensus}
\label{rf-gptoss}
\end{figure}

Figure~\ref{rf-gptoss} illustrates the consolidated reasoning output produced by OpenAI-gpt-oss alongside the individual VLM predictions. The reasoning agent amplifies strong multi-model agreement, resolves conflicting classifications, and explicitly downgrades confidence in ambiguous cases. RF signals consistently classified as \textit{Unknown} across multiple models are assigned high-confidence anomalous labels, while cases with divergent interpretations are flagged for cautious handling or further analysis. This behavior reflects responsible uncertainty management rather than overconfident automation.

This use case demonstrates how the proposed architecture operationalizes Responsible and Explainable AI principles for RF-layer security in 5G networks. Explainability is achieved through parallel multi-model RF interpretation that surfaces uncertainty, disagreement, and alternative classifications at the signal level. Responsibility is enforced through centralized reasoning-layer governance that constrains decision-making, mitigates single-model misclassification, and produces auditable, evidence-backed security judgments~\cite{explainable-ai-text, responsible-llm}. Compared to traditional packet-level intrusion detection systems and single-model RF classifiers, the consensus-driven Deep-RF workflow improves detection robustness, reduces false confidence, and strengthens operational trust, providing a practical foundation for next-generation AI-driven 5G security monitoring.

\section{Related Work}

\begin{table*}[!htb]
\centering
\vspace{0.1in}
\caption{Comparison of Related Multi-Model and Agentic AI Systems with the Proposed Architecture}
\label{t_agentic_comparison}
\begin{adjustbox}{width=1\textwidth}
\begin{tabular}{lcccccc}
\toprule
\thead{Reference / System} &
\thead{Supports\\RAI} &
\thead{Supports\\XAI} &
\thead{Multi-Model\\Consensus} &
\thead{Reasoning\\LLM Layer} &
\thead{Supported\\LLMs / Models} &
\thead{Evaluated\\Use Cases} \\
\midrule

\textbf{This Work} 
& \cmark 
& \cmark 
& \cmark 
& \cmark 
& \makecell{LLM-agnostic\\GPT, Gemini, Claude,\\ Llama, Pixtral, Qwen} 
& \makecell{Agentic AI Workflows\\ Podcast generation,\\ H-reflex analysis,\\ Dental imaging,\\ RF signal classification,\\ Psychiatric diagnosis} \\

\midrule

LLM Ensemble~\cite{llm-ensemble}
& \xmark 
& \pmark 
& \cmark 
& \xmark 
& \makecell{GPT, LLaMA, PaLM,\\ Mixtral (surveyed)} 
& \makecell{General NLP tasks\\ (QA, reasoning, coding)} \\

Reconcile~\cite{reconcile}
& \xmark 
& \pmark 
& \cmark 
& \xmark 
& \makecell{GPT-3.5, GPT-4} 
& \makecell{Reasoning benchmarks,\\ math, QA} \\

Reliable Multi Agent~\cite{reliabl-multi-agent}
& \pmark 
& \pmark 
& \cmark 
& \xmark 
& \makecell{GPT-family,\\ proprietary LLMs} 
& \makecell{Decision-making\\ simulations} \\

Agentic Robotics~\cite{agentic-robotics}
& \xmark 
& \pmark 
& \cmark 
& \xmark 
& \makecell{GPT-based LLMs} 
& \makecell{Robotics reasoning\\ and planning} \\

Governance-as-a-Service~\cite{governance-as-service}
& \cmark 
& \pmark 
& \xmark 
& \cmark 
& \makecell{Policy engines +\\ LLM backends} 
& \makecell{AI governance,\\ compliance workflows} \\

Production Agentic Workflows~\cite{agentic-workflow-practicle-guide}
& \cmark 
& \cmark 
& \cmark 
& \cmark 
& \makecell{LLM-agnostic\\ (Workflow Focus)} 
& \makecell{Agentic AI Workflows} \\

\bottomrule
\end{tabular}
\end{adjustbox}
\end{table*}

Recent advances in LLMs and VLMs have led to the rapid adoption of AI systems across diverse domains, including content generation, biomedical analysis, signal processing, and clinical decision support. While many of these systems demonstrate strong task-level performance, most rely on single-model inference pipelines and lack explicit architectural mechanisms for explainability, responsibility, and governance~\cite{reliabl-multi-agent}. As agentic AI systems increasingly operate autonomously and interact with downstream systems, the absence of multi-model transparency and centralized reasoning control poses significant risks related to hallucinations, bias amplification, and untraceable decisions.

A growing body of work has explored ensemble learning and multi-model collaboration to improve robustness and accuracy. However, existing approaches typically focus on voting-based aggregation or heuristic reconciliation, without introducing a dedicated reasoning layer responsible for structured consolidation, conflict resolution, and policy enforcement~\cite{llm-explainability}. In contrast, our work introduces a generalizable agent architecture that integrates an LLM/VLM consortium with a reasoning-layer governance agent, explicitly separating decision generation from decision arbitration. This section reviews representative related work and highlights the architectural gaps addressed by our proposed system.

\subsection{Survey of Large Language Model Ensembles}
Chen et al.\ presented a comprehensive survey of ensemble techniques for large language models, analyzing methods such as majority voting, weighted averaging, and confidence-based selection across multiple LLMs~\cite{llm-ensemble}. Their study demonstrates that ensemble strategies can improve robustness and reduce variance in tasks such as question answering, reasoning, and code generation. However, the surveyed approaches treat aggregation as a statistical or heuristic post-processing step and do not incorporate a dedicated reasoning agent capable of structured comparison, justification, or policy-aware governance. As such, while ensemble diversity is leveraged for accuracy, explainability and responsibility remain implicit rather than architecturally enforced.

\subsection{Reconciliation of Multiple LLM Outputs for Reasoning Tasks}
Chen et al.\ introduced ReConcile, a framework that combines outputs from multiple LLMs to improve reasoning performance on arithmetic and logical benchmarks~\cite{reconcile}. The system identifies overlapping reasoning steps and resolves inconsistencies through iterative refinement. Although ReConcile demonstrates improved accuracy over single-model baselines, it does not expose intermediate outputs as first-class artifacts for inspection, nor does it provide explicit mechanisms for uncertainty signaling or downstream governance. The reconciliation process is tightly coupled to task-specific heuristics rather than a reusable reasoning-layer abstraction.

\subsection{Reliable Multi-Agent Decision-Making with Large Language Models}
Lee et al.\ proposed a multi-agent framework in which several LLM agents collaborate to solve complex decision-making problems~\cite{reliabl-multi-agent}. Their work emphasizes redundancy and cross-verification among agents to reduce erroneous decisions. While the framework improves reliability compared to single-agent systems, it relies primarily on peer discussion and convergence dynamics, without introducing a centralized reasoning authority to enforce constraints, resolve conflicts deterministically, or ensure auditability. As a result, responsibility and explainability are emergent properties rather than guaranteed architectural features.

\subsection{Agentic LLM Architectures for Robotics and Planning}
Moncada-Ramirez et al.\ explored agentic LLM architectures for robotics planning and control, where multiple agents collaborate to generate action plans and task decompositions~\cite{agentic-robotics}. Their approach demonstrates the feasibility of distributed reasoning across agents in embodied environments. However, the system focuses primarily on task completion efficiency and does not explicitly address responsible AI concerns such as traceability, policy compliance, or explainable consolidation of competing plans. Model outputs are often merged implicitly through dialogue rather than governed through a formal reasoning layer.

\subsection{Governance-as-a-Service for AI Systems}
Recent work on Governance-as-a-Service frameworks introduces centralized policy engines and compliance layers to monitor and constrain AI behavior~\cite{governance-as-service}. These systems emphasize regulatory alignment, audit logging, and risk management, particularly in enterprise and regulated environments. While such frameworks provide strong responsibility guarantees, they typically operate independently of model-level reasoning and do not leverage multi-model consensus for explainability. Consequently, governance is enforced externally rather than emerging from transparent, model-driven comparison and reasoning.

\subsection{Production-Oriented Agentic AI Systems}
Bandara et al.\ discussed architectural challenges and best practices for deploying agentic AI systems in production environments~\cite{agentic-workflow-practicle-guide}. Their work highlights issues such as orchestration, scalability, and fault tolerance, emphasizing the need for modular agent design. However, the proposed architectures remain largely model-agnostic and do not address how explainability or responsibility can be systematically embedded into agent decision pipelines. Multi-model reasoning and explicit governance layers are identified as open challenges rather than resolved components.

\subsection{Positioning of the Proposed Architecture}

Table~\ref{t_agentic_comparison} presents a comparative analysis of existing AI-based dental diagnostic frameworks across several key dimensions, including fine-tuning adaptability, runtime integration of VLMs or LLMs, vision-language modeling capabilities, reasoning LLM utilization, and multi-model orchestration support. 

Table~\ref{t_agentic_comparison} presents a comparative analysis of existing AI systems and frameworks with respect to their explicit support for Responsible AI (RAI) and Explainable AI (XAI) principles, with particular emphasis on multi-model and agentic AI capabilities. In contrast to prior work, the proposed architecture unifies multi-model consensus execution and reasoning-layer governance into a single, reusable agentic pattern that enforces explainability and responsibility by design rather than as post-hoc additions~\cite{explainable-ai-text, responsible-llm}. By preserving independent outputs from heterogeneous LLMs and VLMs and consolidating them through a dedicated reasoning agent, the system provides explainability through observable disagreement and responsibility through centralized control. Unlike ensemble or dialogue-based approaches, decision governance is explicit, auditable, and domain-agnostic, enabling deployment across heterogeneous use cases including neuromuscular reflex analysis, dental imaging, RF signal classification, and psychiatric decision support. This architectural separation of generation and governance distinguishes our work from existing multi-agent and ensemble-based AI systems.

\section{Conclusions and Future Work}

This paper presented a consensus-driven Responsible and Explainable Agent Architecture for designing, deploying, and governing production-grade agentic AI systems. The proposed architecture explicitly separates task execution from decision governance by combining parallel multi-model inference with a centralized reasoning layer, enabling robust, transparent, and accountable AI behavior across diverse application domains. We demonstrated the generality and effectiveness of the architecture through five real-world agentic AI workflows: news podcast generation, neuromuscular reflex analysis, detection of dental conditions and gingivitis, psychiatric diagnosis, and classification of RF signals. These use cases span the domains of content generation, biomedical signal analysis, clinical decision support, and security monitoring, characterized by varying levels of risk, uncertainty, and accountability requirements. Across all workflows, the architecture consistently improved robustness by mitigating single-model failure modes, enhanced explainability by exposing agreement and disagreement across models, and strengthened responsibility through reasoning-layer governance that filtered speculative outputs and enforced domain-aligned constraints. Rather than optimizing solely for task-level accuracy, our evaluation emphasized properties essential to Responsible and Explainable AI, including uncertainty surfacing, traceable reasoning, reproducibility, and auditability. The results show that consensus-driven reasoning enables more reliable and trustworthy AI-assisted decision support compared to traditional single-model pipelines, particularly in high-stakes environments where overconfidence and opaque automation pose significant risks. By embedding responsibility and explainability directly into agent design and orchestration, this work contributes toward the development of trustworthy, production-ready agentic AI systems capable of safe and reliable deployment across critical real-world domains. For future work, we plan to extend the proposed architecture to a broader set of agentic AI workflow–automation use cases, validating their effectiveness across diverse domains and increasingly complex multi-agent pipelines.



\bibliographystyle{elsarticle-num}
\bibliography{reference}

\end{document}